\documentclass[letterpaper]{article} 
\usepackage{aaai2026}  
\usepackage{times}  
\usepackage{helvet}  
\usepackage{courier}  
\usepackage[hyphens]{url}  
\usepackage{graphicx} 
\usepackage{natbib}  
\usepackage{caption} 
\usepackage{algorithm}
\usepackage{algorithmic}

\usepackage{amsfonts}
\usepackage{amsmath}
\usepackage{subcaption}

\usepackage{newfloat}
\usepackage{listings}
\DeclareCaptionStyle{ruled}{labelfont=normalfont,labelsep=colon,strut=off} 
\floatstyle{ruled}
\newfloat{listing}{tb}{lst}{}
\floatname{listing}{Listing}
\title{PhysicsCorrect: A Training-Free Approach for Stable Neural PDE Simulations}
\author{
    Xinquan Huang\textsuperscript{\rm 1}\thanks{Corresponding author}, 
    Paris Perdikaris\textsuperscript{\rm 1}
}
\affiliations{
    \textsuperscript{\rm 1}University of Pennsylvania\\
    \{huang26, pgp\}@seas.upenn.edu
}

\begin{document}

\maketitle

\begin{abstract}
Neural networks have emerged as powerful surrogates for solving partial differential equations (PDEs), offering significant computational speedups over traditional methods. However, these models suffer from a critical limitation: error accumulation during long-term rollouts, where small inaccuracies compound exponentially, eventually causing complete divergence from physically valid solutions. We present PhysicsCorrect, a training-free correction framework that enforces PDE consistency at each prediction step by formulating correction as a linearized inverse problem based on PDE residuals. Our key innovation is an efficient caching strategy that precomputes the Jacobian and its pseudoinverse during an offline warm-up phase, reducing computational overhead by two orders of magnitude compared to standard correction approaches. Across three representative PDE systems, including Navier-Stokes fluid dynamics, wave equations, and the chaotic Kuramoto-Sivashinsky equation, PhysicsCorrect reduces prediction errors by up to 100× while adding negligible inference time (under 5\%). The framework integrates seamlessly with diverse architectures, including Fourier Neural Operators, UNets, and Vision Transformers, effectively transforming unstable neural surrogates into reliable simulation tools that bridge the gap between deep learning's computational efficiency and the physical fidelity demanded by practical scientific applications. 
The code is available at\\ \url{https://github.com/summerwine668/PhysicsCorrect}. 
\end{abstract}


\section{Introduction}
Simulating physical systems governed by partial differential equations (PDEs) is fundamental to numerous scientific and engineering disciplines. Achieving stable long-term rollouts is especially critical for applications such as optimal control, inverse design, and computational imaging. While classical numerical methods like finite-difference \citep{smith1985numerical}, finite-element \citep{reddy1993introduction}, and spectral-element methods \citep{patera1984spectral} provide accurate solutions, they often demand substantial computational resources, limiting their real-time applicability.

Neural PDE solvers have emerged as promising alternatives offering significant computational efficiency \citep{guo2016convolutional,zhu2019physics,geneva2020modeling,wandel2021teaching,huang2024lordnet,zhou2024unisolver,alkin2024universal,gao2025generative}. These approaches approximate PDE solution operators, enabling rapid inference once trained. However, they face a fundamental challenge: error accumulation during autoregressive rollouts, where small errors compound exponentially, leading to numerical instability and divergence \citep{Sanchez-Gonzalez2020}.

\begin{figure}
\centering
\includegraphics[width=1.0\linewidth]{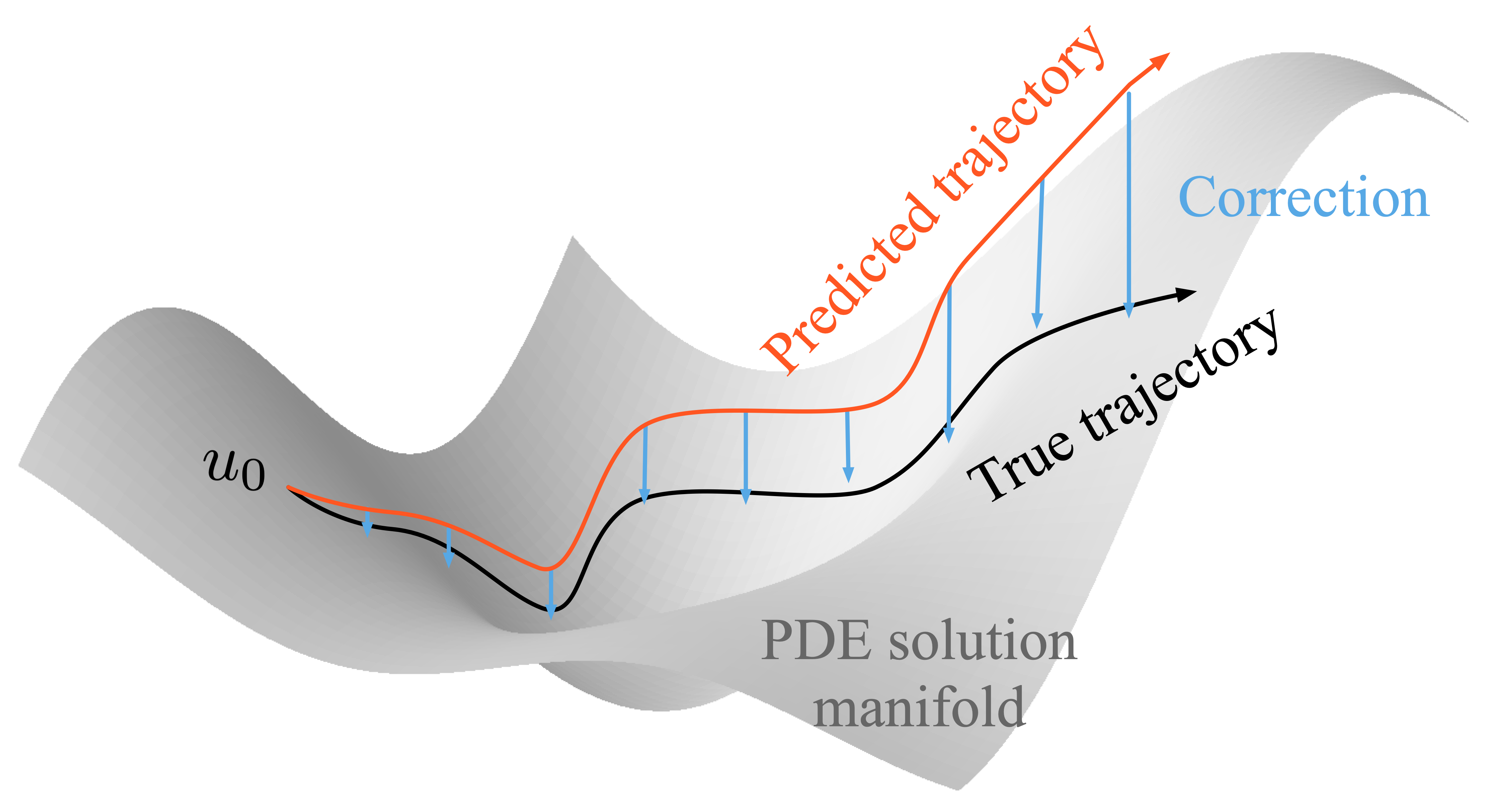}
\caption{PhysicsCorrect stabilizes neural PDE solver rollouts by projecting erroneous predictions back onto the manifold of physically consistent solutions.}
\label{fig:pde_manifold}
\end{figure}
Current mitigation strategies, such as injecting random noise during training \citep{Sanchez-Gonzalez2020} or multi-step training regimes, often fall short because prediction errors are structured rather than random. Newer approaches like PDE-refiner \citep{lippe_pde-refiner_2023} employ generative diffusion models at each step, but introduce substantial computational overhead that may negate the speed advantages of neural solvers.

We propose a fundamentally different approach that directly corrects each prediction using the governing PDE itself. Our key insight is that the PDE residual provides a natural signal for correction. By formulating this as a linear inverse problem based on the Jacobian of the PDE residual, we efficiently project predictions onto the manifold of physically consistent solutions (Figure~\ref{fig:pde_manifold}).

Our approach requires no additional training, operates efficiently during inference, and is compatible with any pretrained neural PDE solver. For many PDEs, the Jacobian matrix and its pseudoinverse can be precomputed in an offline warm-up phase, resulting in minimal computational overhead. Even for highly nonlinear PDEs with imperfect Jacobian approximations, the correction significantly improves long-term stability. Our contributions are:
\begin{itemize}
    \item A physics-informed correction framework that leverages PDE residuals for stable long-term rollouts without retraining neural models.
    \item An efficient caching strategy that precomputes the Jacobian pseudoinverse during an offline phase, reducing computational burden while maintaining accuracy.
    \item Demonstration of broad applicability across multiple PDE types (Navier-Stokes, wave equation, Kuramoto-Sivashinsky) and neural architectures (FNO, UNet, ViT), showing consistent improvements in accuracy and stability.
\end{itemize}
 
In the following section, we review related work on neural PDE solvers and existing approaches for improving rollout stability, before detailing our physics-informed correction framework in Section \ref{sec:method} and validating its performance across diverse PDE systems in Section \ref{sec:numerical}.

\section{Background \& Related Work}
\label{sec:related_work}

\paragraph{Neural PDE Solvers.}
Neural operators have emerged as powerful surrogates for solving time-dependent PDEs, offering significant computational speedups over traditional numerical methods by learning mappings between function spaces to approximate PDE solution operators.

The Fourier Neural Operator (FNO) \citep{li_fourier_2020} established a foundation by performing spectral domain convolutions to capture global dependencies with excellent generalization. The field has since evolved with specialized architectures: message-passing networks \citep{brandstetter_message_2022} for complex geometries, Clifford neural networks \citep{brandstetter_clifford_2022} for physical invariances, and transformer-based approaches like GNOT \citep{hao_gnot_2023}, Transolver \citep{wu_transolver_2024}, and CViT \citep{wang_cvit_2025} that combine attention mechanisms with physical priors.

For time-dependent PDEs, these models are typically applied autoregressively, predicting the next state from the current one. While they excel at one-step predictions on in-distribution data, they struggle with long-term rollout stability. Small errors accumulate and amplify over multiple time steps, eventually causing catastrophic divergence, one of the most significant barriers to deploying neural operators in real-world applications.

\paragraph{Strategies for Improving Rollout Stability.}
Several approaches have been developed to address this challenge. Sanchez-Gonzalez \textit{et al.} \cite{Sanchez-Gonzalez2020} introduced adversarial training with random noise injection to build robustness against perturbations, while Brandstetter \textit{et al.} \cite{brandstetter_message_2022} proposed multi-step training where loss is computed over prediction sequences, allowing models to compensate for their errors.

Recent advances leverage generative models: PDE-Refiner \citep{lippe_pde-refiner_2023} adapts diffusion models to iteratively denoise predicted states, and Hu \textit{et al.} \cite{hu_wavelet_2024} employs wavelet-based diffusion to improve spectral accuracy. While effective, these methods introduce substantial computational overhead during training or inference, often negating the efficiency advantages of neural surrogates.

\paragraph{Physics-Based Correction Approaches.}
An alternative strategy leverages the governing equations to correct predictions without requiring model retraining. Cao \textit{et al.} \cite{cao_residual-based_2023} developed PDE residual minimization methods for Bayesian inference using goal-oriented a-posteriori error estimation \citep{jha_goal-oriented_2022}, later extended to nonlinear variational problems \citep{jha_residual-based_2024}. While promising, these approaches often introduce significant computational overhead.

Other research has focused on enforcing specific physical constraints. Jiang \textit{et al.} \cite{jiang_enforcing_2020} implemented spectral projection layers for divergence-free conditions in fluid simulations, and Duruisseaux \textit{et al.} \cite{duruisseaux_towards_2024} generalized this to broader linear differential constraints. These methods are computationally efficient but limited to specific physical aspects rather than ensuring complete PDE satisfaction.

\paragraph{Our Approach.}
Our work bridges these approaches through a correction framework that enforces PDE satisfaction at each timestep via residual minimization. Unlike existing methods, our approach requires no additional training, employs efficient caching strategies to minimize computational overhead, and generalizes across different PDE types and neural architectures. By treating physical consistency as an online correction problem, we achieve stable long-term rollouts while preserving the computational advantages of neural PDE solvers.

\section{Methodology}
\label{sec:method}

\paragraph{Long-term Evolution with Neural PDE Solvers.}
Consider time-dependent partial differential equations of the general form:
\begin{equation}
   \label{equ:governpde}
   S\left(\mathbf{u}, \frac{\partial\mathbf{u}}{\partial t}, \frac{\partial^2\mathbf{u}}{\partial^2 t},..., \frac{\partial\mathbf{u}}{\partial \mathbf{x}}, \frac{\partial^2\mathbf{u}}{\partial^2 \mathbf{x}},...\right) = 0,
\end{equation}
where $\mathbf{u}$ represents the solution defined on spatial domain $\mathbf{x}\in\mathcal{X}$ and temporal domain $t\in [0, T]$.

Neural PDE solvers approximate the time-evolution operator of such systems. Given a current state $\mathbf{u}(\mathbf{x},t)$, a neural network $\phi_\theta$ with parameters $\theta$ predicts the state at the next time step:
\begin{equation}
   \label{equ:timepde}
   \mathbf{u}(\mathbf{x},t+\Delta t)=\phi_\theta(\mathbf{u}(\mathbf{x},t)).
\end{equation}

For long-term simulations starting from an initial condition $\mathbf{u}_0$, the state at time $t$ is obtained through repeated application of $\phi_\theta$:
\begin{equation}
   \label{equ:longrollout}
   \mathbf{u}_t = \phi_\theta(\phi_\theta(...\phi_\theta(\mathbf{u}_{0}) + \epsilon_{0}...)+\epsilon_{t-1}),
\end{equation}
where $\epsilon_t$ represents the prediction error at step $t$.

\paragraph{The Challenge of Error Accumulation.}
The fundamental challenge in autoregressive rollouts is that prediction errors compound over time, often leading to numerical instability or complete divergence from the true solution trajectory. Existing approaches attempt to address this problem by either: (1) \textit{Enhancing model robustness} through specialized training regimes like noise injection or multi-step training, which requires substantial additional training data and computational resources; or (2) \textit{Applying post-hoc corrections} through computationally expensive denoising procedures that may negate the efficiency advantages of neural solvers.

Both approaches face limitations because prediction errors exhibit complex, structured patterns rather than random noise. These errors depend on the specific state distribution and are difficult to anticipate through data augmentation alone. Moreover, even highly accurate neural operators may eventually diverge during sufficiently long rollouts.

Here we propose an alternative paradigm: a lightweight, physics-informed correction mechanism that operates during inference without requiring model retraining. By explicitly minimizing the PDE residual at each time step, we project predictions back onto the manifold of physically valid solutions, effectively transforming a challenging multi-step prediction problem into a sequence of more manageable one-step predictions. This approach aims to strike an optimal balance between computational efficiency and numerical stability, providing a general solution for accurate long-term simulations across different PDE types and neural architectures.

\subsection{The PhysicsCorrect Framework: Linearized PDE Residual Correction}

\paragraph{Problem Formulation.}
The core idea of our PhysicsCorrect approach is to leverage the governing PDE itself as a form of implicit supervision, correcting neural network predictions to better satisfy the underlying physics. For a state $\mathbf{u}_t$ at time $t$, a neural operator produces a prediction $\hat{\mathbf{u}}_{t+1}$ for the next time step. Our goal is to find a correction term $\mathbf{u}_{t+1}^c$ such that the corrected prediction $\hat{\mathbf{u}}_{t+1} + \mathbf{u}_{t+1}^c$ better approximates the true solution $\mathbf{u}_{t+1}$.

While the ground truth $\mathbf{u}_{t+1}$ is unavailable during inference, we can evaluate how well a candidate solution satisfies the governing equation by computing the PDE residual. For a discretized PDE, this residual $L_{\text{PDE}}(\mathbf{u}_t, \mathbf{u}_{t+1})$ is obtained by substituting $\mathbf{u}_t$ and $\mathbf{u}_{t+1}$ into Equation \ref{equ:governpde}. A physically consistent solution would yield a residual of zero (or for second-order time derivatives, the residual would depend on $\mathbf{u}_t$, $\mathbf{u}_{t+1}$, and $\mathbf{u}_{t-1}$).

\paragraph{Efficient Correction via Linear Approximation.}
A natural approach would be to directly minimize the PDE residual with respect to the correction term:
\begin{equation}
   \label{equ:l2pde}
   \mathbf{u}_{t+1}^{c*} = \underset{\mathbf{u}_{t+1}^c}{\arg\min}\; \| L_{\text{PDE}}(\mathbf{u}_t, \hat{\mathbf{u}}_{t+1}+\mathbf{u}_{t+1}^c) \|^2.
\end{equation}
However, directly solving this optimization problem would require iterative methods with adaptive learning rates, introducing significant computational overhead that could negate the efficiency advantages of neural PDE solvers.
Instead, we linearize the problem using a first-order Taylor expansion of the residual around the current prediction $\hat{\mathbf{u}}_{t+1}$:
\begin{align}
    \label{equ:taylor}
   L_{\text{PDE}}(\mathbf{u}_t, \hat{\mathbf{u}}_{t+1}+\mathbf{u}_{t+1}^c) &\approx L_{\text{PDE}}(\mathbf{u}_t, \hat{\mathbf{u}}_{t+1}) 
   \notag\\&+ \frac{\partial L_{\text{PDE}}(\mathbf{u}_t, \hat{\mathbf{u}}_{t+1})}{\partial \hat{\mathbf{u}}_{t+1}}\mathbf{u}_{t+1}^c.
\end{align}
This approximation is valid when the correction term $\mathbf{u}_{t+1}^c$ is sufficiently small, a condition typically
\begin{figure}
    \centering
\includegraphics[width=1.0\linewidth]{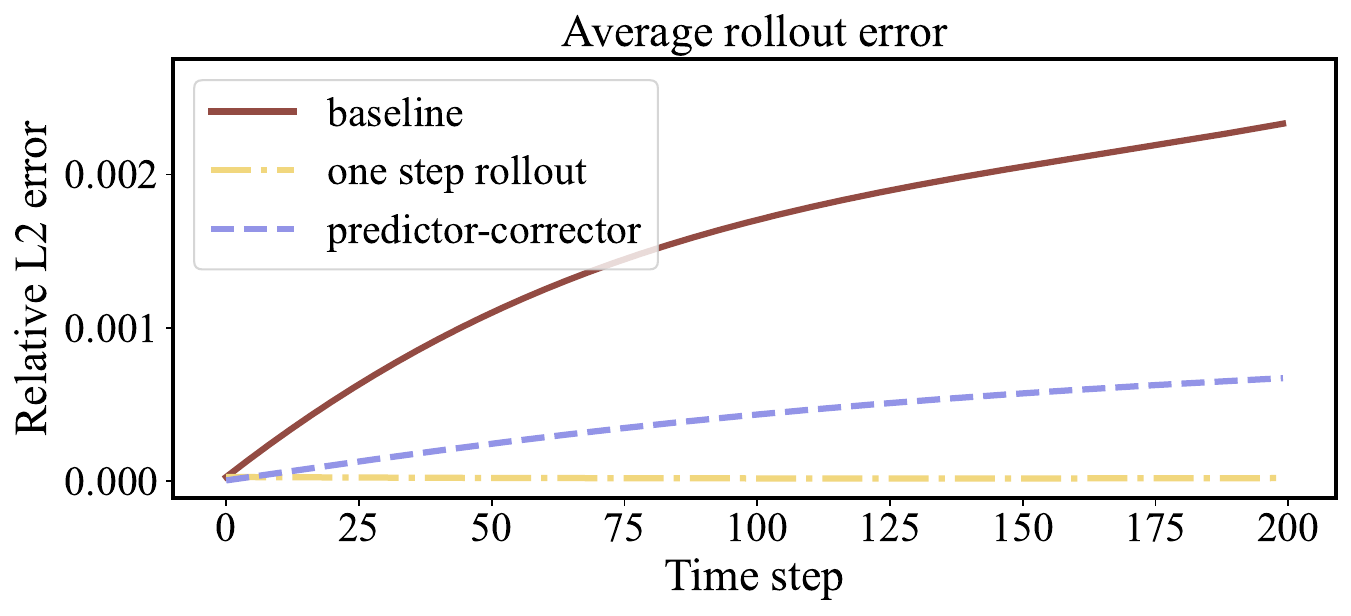}
\caption{Long-term rollout accuracy comparison for the 2D Navier-Stokes benchmark. The baseline neural operator (brown) exhibits error accumulation, while our predictor-corrector approach (blue) maintains stability throughout the simulation, closely matching the performance of idealized one-step rollouts (yellow).}
\label{fig:diagram-nse-sample}
\end{figure}
satisfied when the neural operator has been adequately trained for direct full physical prediction (directly approximate $\mathbf{u}_{t+1}$). When using the residual prediction ($\mathbf{u}_{t+1} - \mathbf{u}_{t}$), this assumption will be relaxed (see the sensitivity analysis in the Appendix). Setting the linearized residual to zero yields a linear system:
\begin{equation}
   \label{equ:axb}
   \frac{\partial L_{\text{PDE}}(\mathbf{u}_t, \hat{\mathbf{u}}_{t+1})}{\partial \hat{\mathbf{u}}_{t+1}}\mathbf{u}_{t+1}^c = -L_{\text{PDE}}(\mathbf{u}_t, \hat{\mathbf{u}}_{t+1}).
\end{equation}
This can be expressed in the standard form $A\mathbf{x} = \mathbf{b}$, where $A$ is the Jacobian matrix of the PDE residual with respect to $\hat{\mathbf{u}}_{t+1}$, $\mathbf{x}$ is the correction term $\mathbf{u}_{t+1}^c$ we seek to determine, and $\mathbf{b}$ is the negative PDE residual $-L_{\text{PDE}}(\mathbf{u}_t, \hat{\mathbf{u}}_{t+1})$.
The correction term can then be obtained by solving this linear system, typically using least-squares methods to handle potential ill-conditioning or over-determined systems. Unlike training-based approaches that learn to predict corrections from residuals, our method directly inverts the linearized PDE operator, avoiding the need for additional training data and mitigating potential distribution shifts between training and inference.

\begin{figure*}
    \centering
    \includegraphics[width=1.\textwidth]{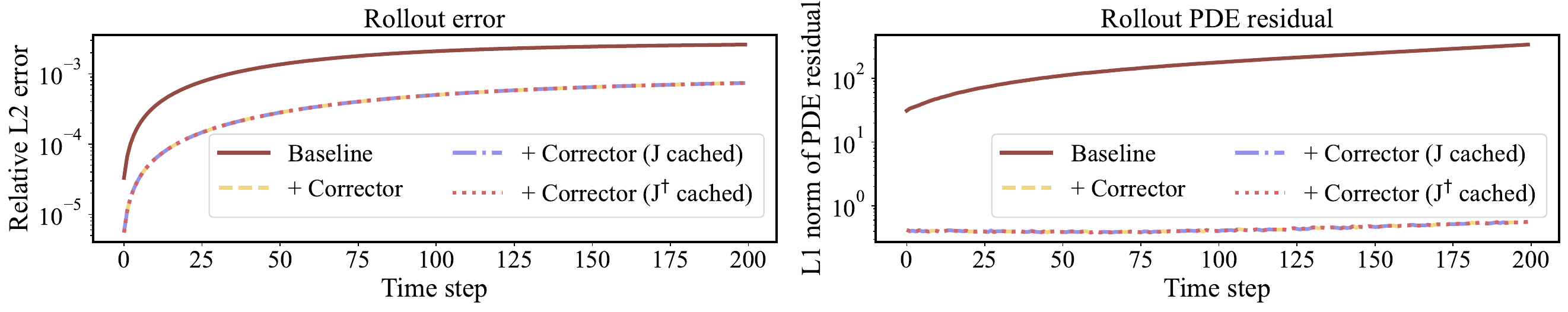}
    \caption{PhysicsCorrect's caching strategy efficiency on 2D Navier-Stokes. Left: Relative $L_2$ error vs. reference solutions over 200 time steps. Right: PDE residual magnitude per step. While the baseline (brown) shows increasing error and residual, both uncached (yellow) and Jacobian-cached (blue) corrections maintain low values. Pseudoinverse caching (red) preserves performance while reducing computational cost by 163x.}
    \label{fig:caching_performance_plot}
\end{figure*}
\paragraph{Predictor-Corrector Pipeline.}
Based on this formulation, we implement PhysicsCorrect as a two-step predictor-corrector pipeline:
\begin{enumerate}
    \item \textbf{Prediction Step}: The neural operator $\phi_\theta$ produces a prediction $\hat{\mathbf{u}}_{t+1} = \phi_\theta(\mathbf{u}_t)$.
    \item \textbf{Correction Step}: We solve the linear system in Equation \ref{equ:axb} to obtain the correction term $\mathbf{u}_{t+1}^c$ and compute the corrected prediction $\tilde{\mathbf{u}}_{t+1} = \hat{\mathbf{u}}_{t+1} + \mathbf{u}_{t+1}^c$.
\end{enumerate}
This corrected state $\tilde{\mathbf{u}}_{t+1}$ then serves as input for the next prediction step. By ensuring that each state 
better satisfies the underlying PDE, we significantly reduce error accumulation during autoregressive rollouts. As shown in Figure \ref{fig:diagram-nse-sample}, this approach maintains stability and accuracy over hundreds of time steps when applied to the 2D Navier-Stokes equation, while the baseline prediction eventually diverges due to accumulated errors. The figure also illustrates how our method's performance closely approaches that of idealized one-step rollouts, effectively transforming a challenging multi-step prediction problem into a sequence of more manageable one-step predictions.

Our formulation assumes that the neural operator's prediction provides a good initial approximation, allowing the linearization to be effective. This assumption generally holds but may be challenged in highly chaotic systems or with poor initialization (see sensitivity analysis in the Appendix). Nevertheless, the effectiveness of our approach across different PDEs and neural architectures is demonstrated empirically in Section \ref{sec:numerical}, even in challenging regimes.

\paragraph{Efficient Implementation via Jacobian Caching.}
While the physics-informed correction significantly improves rollout stability, a naive implementation would introduce substantial computational overhead from two expensive operations at each time step: (1) evaluating the Jacobian matrix, and (2) solving the resulting least-squares problem.
The Jacobian evaluation requires computing gradients between two spatial fields of dimension $M \times N$, where $M$ and $N$ represent the height and width of the domain. This operation, typically performed through automatic differentiation, scales poorly with increasing resolution. Similarly, solving the least-squares problem via singular value decomposition is numerically stable but computationally intensive.

We introduce a key optimization that dramatically reduces this computational burden. For many time-dependent PDEs, we observe that the Jacobian matrix remains constant across different time steps and initial conditions when the residual $L_{\text{PDE}}(\mathbf{u}_t, \hat{\mathbf{u}}_{t+1})$ is linear with respect to $\hat{\mathbf{u}}_{t+1}$. This property allows us to precompute the Jacobian matrix once during an offline warm-up phase, along with its Moore-Penrose pseudoinverse $A^{\dagger}$. During inference, we can then directly compute the correction as $\mathbf{u}^c_{t+1} = A^{\dagger} \mathbf{b}$, where $\mathbf{b} = -L_{\text{PDE}}(\mathbf{u}_t, \hat{\mathbf{u}}_{t+1})$. This approach effectively transforms the correction step from an expensive numerical operation to a simple matrix multiplication, resulting in minimal computational overhead during rollout.

For this caching strategy to be effective, in the Jacobian calculation, the PDE residual must maintain linearity with respect to $\hat{\mathbf{u}}_{t+1}$. To satisfy this requirement while preserving numerical stability, we employ a semi-implicit discretization scheme that treats linear terms (e.g., diffusion) implicitly to ensure stability, while handling nonlinear terms (e.g., advection) explicitly to preserve Jacobian constancy. For example, in the 2D Navier-Stokes equations, we implement a Crank-Nicolson scheme with implicit diffusion and explicit advection terms (more details are in the Appendix). 
This formulation ensures that the Jacobian remains constant across all predictions and initial conditions, while maintaining adequate numerical stability.

As demonstrated in Figure \ref{fig:caching_performance_plot}, our caching strategy achieves accuracy comparable to the non-cached version while reducing computational cost by approximately 160x. The precomputation phase for a 64x64 resolution grid requires only 8.74 seconds, after which the correction adds minimal overhead to the neural operator's inference time (0.90 vs. 0.69 seconds for 200 time steps). 
Importantly, even for chaotic systems where the approximation to the Jacobian matrix is relatively poor with the semi-implicit discretization, the correction still substantially improves rollout stability. We observe that this approach works effectively even when applying it to chaotic systems with nonlinear residual terms, as we demonstrate with the Kuramoto-Sivashinsky equation in Section \ref{sec:numerical:ks}.

\begin{figure*}
    \centering
    \includegraphics[width=1.\linewidth]{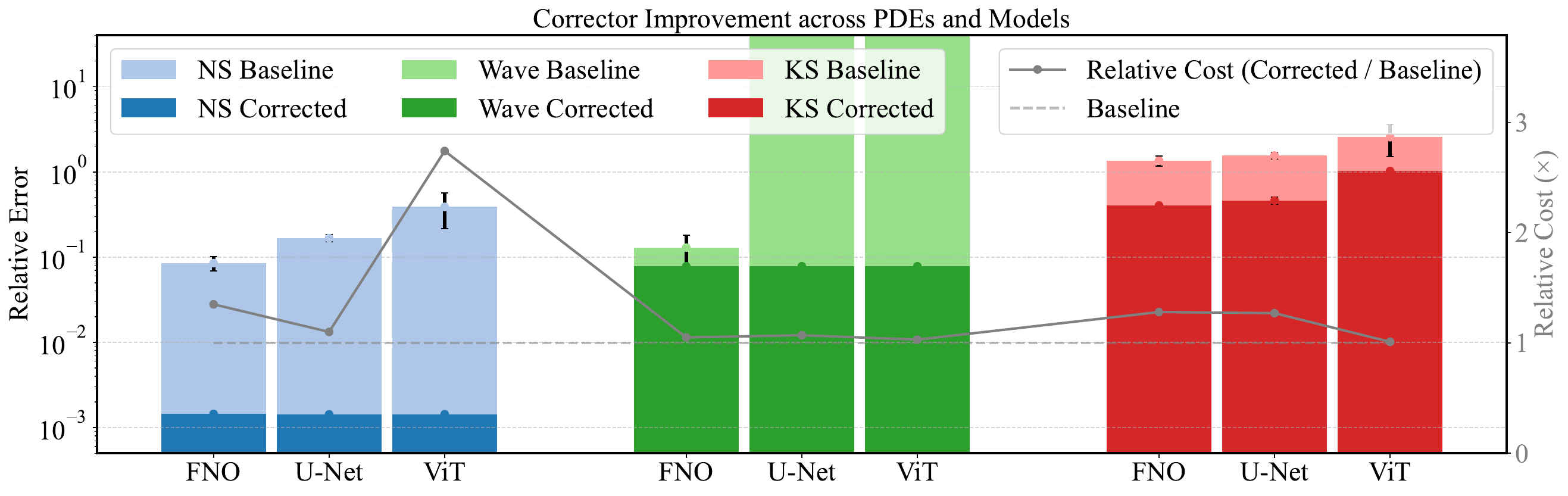}
    \caption{Performance comparison of our physics-informed correction approach across different PDE systems and neural architectures. Left axis (bars): Relative L2 error of baseline models (lighter colors) versus corrected models (darker colors) for Navier-Stokes (NS), wave equation, and Kuramoto-Sivashinsky (KS) equations at the final state after long rollouts (1000 time step rollout for NS and KS equations; 100 time steps for wave equation). 
    Error bars represent the standard deviation for 5 seeds.
    Right axis (line): Relative computational cost of the corrected approach compared to baseline. The correction framework consistently reduces error across all PDEs and architectures with minimal computational overhead. Detailed rollout histories and comparison with baseline models are provided in the Appendix.}
    \label{fig:results_summary}
\end{figure*}

\section{Experiments}
\label{sec:numerical}
We evaluate the PhysicsCorrect framework on three representative PDE systems (Figure~\ref{fig:results_summary}) that vary in complexity, dimensionality, and dynamical behavior: the 2D Navier-Stokes equations (incompressible fluid flow), the 2D wave equation (second-order hyperbolic PDE), and the 1D Kuramoto-Sivashinsky equation (fourth-order nonlinear PDE exhibiting chaotic behavior).
For each system, we test our approach with three neural network architectures: Fourier Neural Operator (FNO) \citep{li_fourier_2020}, UNet \citep{Wandel2020}, and Vision Transformer (ViT) \citep{dosovitskiy_image_2020}. All models are trained to generalize across different initial conditions using the Adam optimizer and L1 loss. Performance is measured by the relative L2 error between predictions and high-fidelity numerical solutions.

Our experiments are designed to address three key questions: whether the correction framework improves the stability and accuracy of long-term rollouts across different neural architectures; how effective the caching strategy is in maintaining accuracy while reducing computational overhead; and how the correction performs on systems with different levels of nonlinearity and chaotic behavior. Network architecture details and training parameters are provided in the Appendix.

\subsection{2D Navier-Stokes equation}
We first evaluate our approach on the 2D incompressible Navier-Stokes equation with a Reynolds number of 1,000 and forcing term $f(x,y)=0.1\sin(2\pi(x+y))+\cos(2\pi(x+y))$ (more details are in the Appendix) 
This system exhibits complex vorticity dynamics and is widely used as a benchmark for fluid simulation.

\paragraph{Experimental Setup.} We train our models on a dataset of 1,000 trajectories generated from Gaussian random initial vorticity fields $\omega_0 \sim \mathcal{N}\left(0,8^{3}(-\Delta+64 I)^{-4.0}\right)$, simulated on a $64\times64$ grid with a time step of 0.01. For the PDE residual formulation, we employ a semi-implicit Crank-Nicolson scheme with explicit advection and implicit diffusion terms. We evaluate performance on 64 test trajectories, each simulated for 1,000 time steps.

\begin{figure}[!htp]
    \centering
    \includegraphics[width=1.0\linewidth]{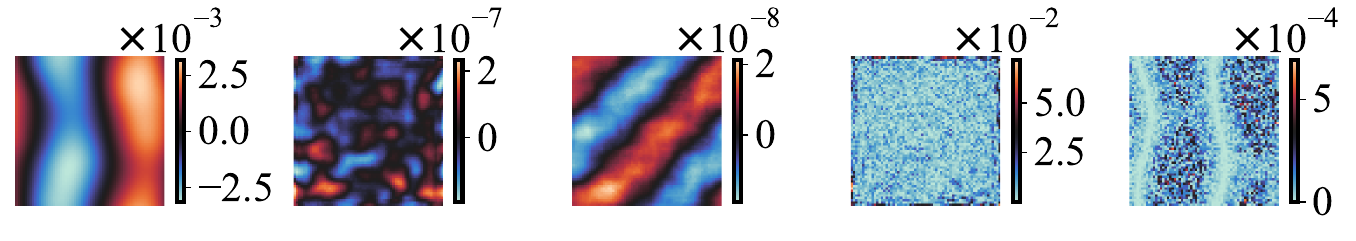}
    \caption{One-step correction on the 2D Navier-Stokes equation. From left to right: ground truth solution from numerical simulation, prediction error of baseline FNO, prediction error after our correction, PDE residual of baseline prediction, and PDE residual after correction. Note the significant reduction in both error magnitude (10× improvement) and PDE residual (100× improvement), demonstrating that our correction effectively projects predictions onto the manifold of physically consistent solutions.}
    \label{fig:corrected-sample-nse}
\end{figure}

\paragraph{One-Step Correction Performance.} Figure~\ref{fig:corrected-sample-nse} demonstrates the effect of our correction on a single prediction step. The baseline FNO prediction contains small but structured errors that our method effectively eliminates. The correction reduces the relative L2 error by an order of magnitude (from 3.3e-5 to 5.5e-6) while simultaneously reducing the PDE residual to near zero, indicating that the corrected state closely satisfies the governing equation.

\paragraph{Long-Term Rollout Stability.} Figure~\ref{fig:results_summary} shows the long-term rollout performance across different neural architectures (FNO, UNet, and ViT). All baseline models exhibit error accumulation that eventually leads to complete divergence from the reference solution. In contrast, models augmented with our physics-informed corrector maintain stable and accurate predictions throughout the entire 1,000-step simulation, regardless of the underlying architecture. The PDE residual (more details and comparison with baseline models are shown in the Appendix)
remains consistently low for all corrected models, confirming that our approach enforces physical consistency at each time step and prevents error accumulation.

The results demonstrate that our correction framework effectively transforms inherently unstable neural PDE solvers into stable simulation tools without requiring architecture-specific modifications or additional training. This universality is particularly valuable as it allows practitioners to leverage any pre-trained neural operator while ensuring physical consistency and long-term stability.

\begin{figure}[!htp]
    \centering
    \includegraphics[width=1.0\linewidth]{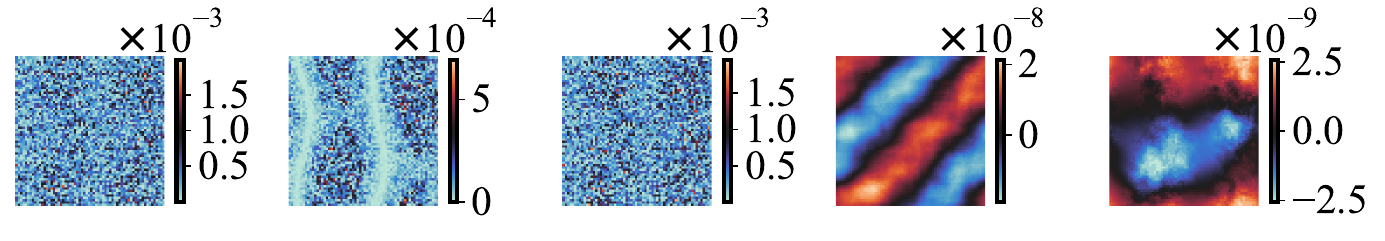}
    \caption{The visualization of one test sample: PDE residual using (from left to right) numerical reference result, corrected prediction, and the corrected prediction with reference PDE residual, the errors of corrected prediction, and the errors of corrected prediction with reference PDE residual.}
    \label{fig:nse_ref_pde}
\end{figure}

\paragraph{Understanding the Limits of Residual-Based Correction.}
Even with perfect numerical methods, discretization introduces small but non-zero PDE residuals in reference solutions, as shown in the leftmost panel of Figure~\ref{fig:nse_ref_pde}. Since our method targets zero residual, this creates a fundamental discrepancy -- we optimize toward a slightly different objective than the true numerical solution, introducing small inherent errors (fourth panel of Figure~\ref{fig:nse_ref_pde}).

To investigate this limitation, we conducted an idealized experiment where we subtract the reference solution's residual from the prediction's residual before correction. This adjustment aligns our optimization target precisely with the numerical reference, resulting in significantly improved accuracy (relative L2 error reduced from 3.3e-5 to 6.1e-7) as shown in the rightmost panel of Figure~\ref{fig:nse_ref_pde}. While this approach is impractical for real applications where reference solutions are unavailable, it demonstrates the theoretical upper bound of our method's performance.

This experiment yields two important insights: first, the quality of PDE discretization directly impacts correction accuracy; and second, even with standard discretization, our method achieves substantial error reduction (approximately 85\%) relative to the theoretical optimum. These findings suggest that using finer discretization schemes for residual computation could further improve correction performance in practice (further discussion is in the Appendix).

\subsection{2D wave equation}
We next evaluate our approach on the 2D wave equation, a second-order linear PDE that models various physical phenomena, including mechanical waves, electromagnetic waves, and seismic propagation (more details are in the Appendix).

\paragraph{Experimental Setup.} We generate data on a $128\times128$ grid using 512 Gaussian random fields as initial conditions with periodic boundaries. Training data consists of the first 10 time steps (recorded at intervals of $\Delta t = 10^{-2}$), while testing evaluates generalization over 100 time steps. For residual computation, we employ an implicit scheme with central finite differences for the time derivatives.

\paragraph{Neural Architecture Considerations.} An interesting finding emerged during our wave equation experiments: standard first-order residual prediction (predicting $\mathbf{u}_{t+1} - \mathbf{u}_t$) consistently failed to capture the essential physics (see Appendix for more results). To resolve this, we revisited the problem through the lens of second-order dynamics, training networks to predict $\delta\mathbf{u}_t = \mathbf{u}_{t+1} + \mathbf{u}_{t-1} - 2\mathbf{u}_t$ instead. This formulation resonates with the oscillatory nature of wave phenomena, providing a significantly more stable formulation. This insight highlights how the representation of physical dynamics can fundamentally shape neural network performance, even before correction mechanisms are applied.

\paragraph{Results.} Figure~\ref{fig:results_summary} shows the long-term rollout performance of different neural architectures with and without our correction framework. Even with the improved second-order formulation, baseline models (particularly ViT and UNet) still exhibit error growth over time. Our physics-informed corrector consistently enhances prediction accuracy across all architectures, maintaining low PDE residuals throughout the simulation. The wave equation results demonstrate our method's effectiveness on linear PDEs with oscillatory dynamics. Interestingly, the performance improvement is more pronounced for architectures that struggle more with the baseline formulation (ViT and UNet), suggesting that our correction approach can help compensate for architecture-specific weaknesses in capturing certain physical dynamics.

\subsection{Kuramoto-Sivashinsky equation}
\label{sec:numerical:ks}

Our final and most challenging test case is the Kuramoto–Sivashinsky (KS) equation, a fourth-order nonlinear PDE that exhibits chaotic dynamics (more details are in the Appendix). 
This system tests our method's effectiveness on strongly nonlinear, chaotic dynamics where prediction errors can amplify rapidly and where linearization approximations are most severely challenged.

\paragraph{Experimental Setup.} We simulate the KS equation on a spatial domain $[0,64]$ with a resolution of 512 points, focusing on the chaotic regime. Starting from $v(x,50)$, we generate 512 trajectories for training (first 500 steps) and 64 trajectories for testing (full 1000 steps) using a spectral method with temporal step size 0.05 \citep{dresdner_learning_2023}.

\paragraph{Challenges with Chaotic Systems.} The KS equation presents unique challenges for our correction approach. The standard strategy of using semi-implicit discretization (implicit for linear terms, explicit for nonlinear terms) to obtain a constant Jacobian proves inadequate due to the equation's strong nonlinearity and chaotic behavior. A fully implicit discretization would provide better residual definition but would require re-computing the Jacobian and its pseudoinverse at each time step, negating the computational advantages of our caching strategy.

\begin{figure*}
    \centering
    \includegraphics[width=\textwidth]{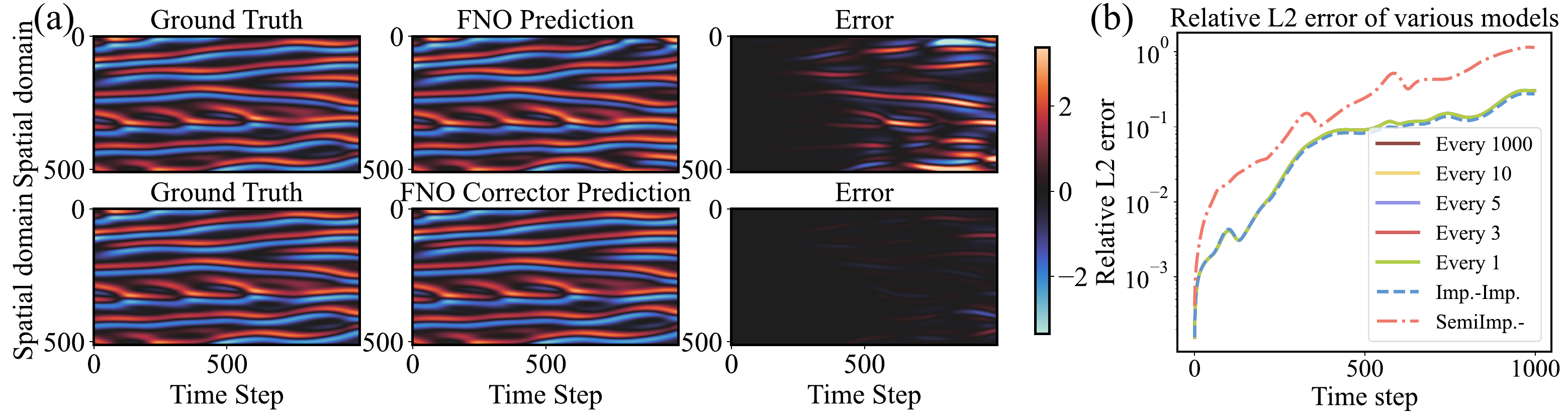}
    \caption{Performance of PhysicsCorrect on the chaotic Kuramoto-Sivashinsky equation. (a) Spatiotemporal evolution over 1000 time steps comparing ground truth, baseline FNO prediction, and error magnitude for uncorrected (top) and corrected (bottom) predictions. (b) Impact of Jacobian update frequency on KS equation prediction error over 1000 steps. Periodic recomputation offers minimal improvement over the fully cached approach, while semi-implicit discretization for both Jacobian and residual shows notably higher errors.}
    \label{fig:ks_comparison}
\end{figure*}
\paragraph{Effectiveness of Approximate Correction.} Surprisingly, as shown in Figure~\ref{fig:results_summary}, our approach with cached pseudoinverse still provides significant stability improvements despite using an approximation of the true Jacobian. Figure~\ref{fig:ks_comparison}(a) shows a qualitative comparison between baseline FNO predictions (top) and corrected predictions (bottom), demonstrating that our method successfully maintains the complex spatiotemporal patterns of the KS equation over long rollouts. 
To further investigate the impact of Jacobian approximation in chaotic systems, we conducted additional experiments varying the frequency of Jacobian updates, as shown in Figure~\ref{fig:ks_comparison}(b). The results reveal that recalculating the Jacobian every 3-10 time steps provides minimal improvement over the fully cached approach (updating only once at initialization). This suggests that for the KS equation, the primary benefit comes from the initial projection toward the physically consistent manifold rather than from having a perfectly accurate Jacobian at each step.

These findings lead to an important insight: accurate definition of the PDE residual is more critical than perfect estimation of the Jacobian pseudoinverse. 
This, in turn, supports the validity of separating the residual evaluation from the Jacobian precomputation, as the correction remains effective even when the precomputed Jacobian is only approximate.
Even with an approximate linear correction, projecting predictions toward the physically consistent manifold provides sufficient regularization to prevent error accumulation. This observation is particularly significant for chaotic systems, where traditional methods often struggle to maintain long-term stability, and offers a favorable trade-off between computational efficiency and prediction accuracy.

\section{Discussion}
\label{sec:discussion}

\paragraph{Summary of Contributions.}
We introduced PhysicsCorrect, a training-free, physics-informed correction framework that significantly enhances neural PDE solver stability during long-term rollouts. Our approach works with any pretrained neural operator with minimal computational overhead through efficient caching. Across Navier-Stokes, wave equation, and Kuramoto-Sivashinsky systems, our method reduces prediction errors by 1-2 orders of magnitude compared to baselines, with consistent benefits across FNO, UNet, and ViT architectures. By enforcing physical consistency at each step, we transform multi-step rollouts into sequences of well-conditioned one-step predictions. Remarkably, even with approximate linearization and cached Jacobians, the method substantially improves stability in chaotic systems, demonstrating robust practical utility.

\begin{figure}[!htp]
    \centering
    \includegraphics[width=1.0\linewidth]{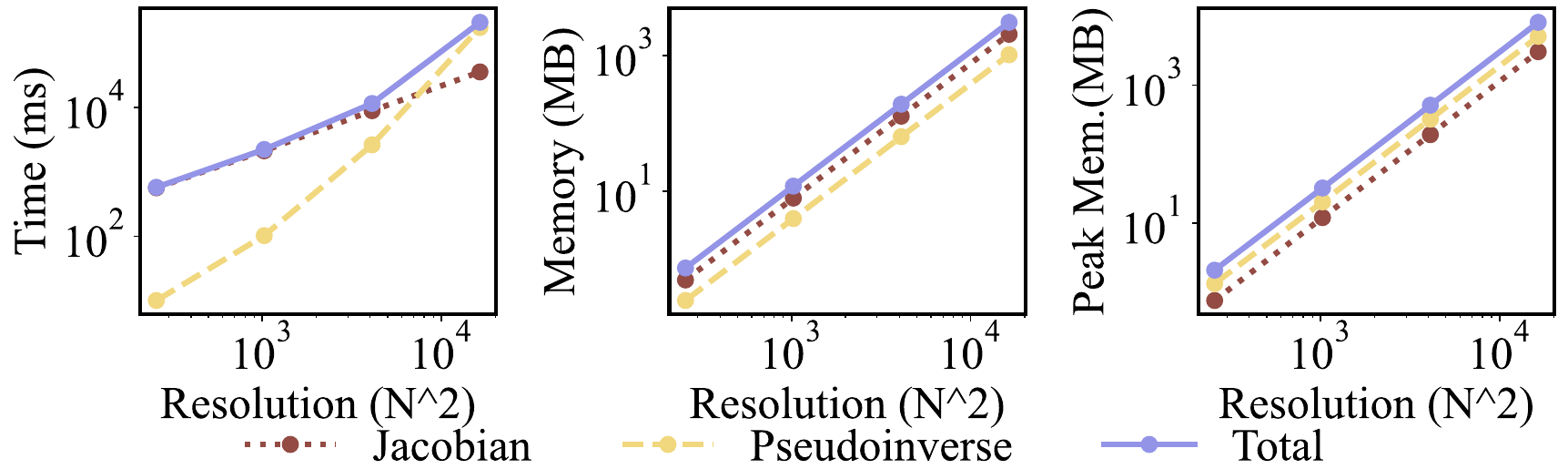}
    \caption{PhysicsCorrect's computational scaling with grid resolution for 2D Navier-Stokes. 
    Plots show computation time (left), GPU memory usage (center), and peak memory consumption (right). All metrics scale quadratically with resolution, highlighting memory and compute requirements as the primary limitation for high-resolution applications, and indicating challenges in scaling to fully 3D problems.}
    \label{fig:scaling_resolution}
\end{figure}
\paragraph{Limitations \& Future Work.}
Despite its effectiveness, PhysicsCorrect faces three key limitations that suggest directions for future work: (1) computational scalability to high-resolution simulations, as Figure~\ref{fig:scaling_resolution} demonstrates quadratic scaling of time and memory requirements with resolution, which could be mitigated by inverting the Jacobian approximately using iterative methods that only require Jacobian-vector products; (2) discretization-induced errors creating an inherent gap between our correction target and true numerical references, potentially improvable through higher-order discretization schemes; moreover, it can be naturally combined with spatio-temporal decomposition training frameworks \citep{huang_neuralstagger_2023}, preserving the large spatial- and temporal-interval advantages of purely neural PDE solvers while enhancing stability through physics-based correction; and (3) potential breakdown of linearization approximations in extreme chaotic systems, which might be addressed by hybrid approaches combining efficient linear correction with occasional nonlinear optimization steps.

\paragraph{Conclusions.}
This work demonstrates that enforcing physical consistency through direct projection onto the manifold of valid solutions provides a powerful, computation-efficient approach to stabilizing neural PDE solvers without requiring additional training or expensive denoising. The method's simplicity, generality, and efficiency make it immediately applicable across scientific and engineering applications, effectively bridging the gap between deep learning's computational advantages and the physical fidelity demanded by practical applications.

\section*{Acknowledgments}
This work was supported by the U.S. Department of Energy, Advanced Scientific Computing Research program, under the “Resolution-invariant deep learning for accelerated propagation of epistemic and aleatory uncertainty in multi-scale energy storage systems, and beyond” project (Project No. 81824), and Penn AIxScience fellowship.

\bibliography{picor}

\begin{thebibliography}{34}
\providecommand{\natexlab}[1]{#1}

\bibitem[{Alkin et~al.(2024)Alkin, F{\"u}rst, Schmid, Gruber, Holzleitner, and Brandstetter}]{alkin2024universal}
Alkin, B.; F{\"u}rst, A.; Schmid, S.; Gruber, L.; Holzleitner, M.; and Brandstetter, J. 2024.
\newblock Universal physics transformers: A framework for efficiently scaling neural operators.
\newblock \emph{Advances in Neural Information Processing Systems}, 37: 25152--25194.

\bibitem[{Brandstetter et~al.(2022)Brandstetter, Berg, Welling, and Gupta}]{brandstetter_clifford_2022}
Brandstetter, J.; Berg, R. v.~d.; Welling, M.; and Gupta, J.~K. 2022.
\newblock Clifford {Neural} {Layers} for {PDE} {Modeling}.
\newblock ArXiv:2209.04934 [physics].

\bibitem[{Brandstetter, Worrall, and Welling(2022)}]{brandstetter_message_2022}
Brandstetter, J.; Worrall, D.; and Welling, M. 2022.
\newblock Message {Passing} {Neural} {PDE} {Solvers}.
\newblock \emph{arXiv}, 1--27.
\newblock ArXiv: 2202.03376.

\bibitem[{Cao et~al.(2023)Cao, O'Leary-Roseberry, Jha, Oden, and Ghattas}]{cao_residual-based_2023}
Cao, L.; O'Leary-Roseberry, T.; Jha, P.~K.; Oden, J.~T.; and Ghattas, O. 2023.
\newblock Residual-based error correction for neural operator accelerated infinite-dimensional {Bayesian} inverse problems.
\newblock \emph{Journal of Computational Physics}, 486: 112104.

\bibitem[{Dosovitskiy et~al.(2020)Dosovitskiy, Beyer, Kolesnikov, Weissenborn, Zhai, Unterthiner, Dehghani, Minderer, Heigold, Gelly, Uszkoreit, and Houlsby}]{dosovitskiy_image_2020}
Dosovitskiy, A.; Beyer, L.; Kolesnikov, A.; Weissenborn, D.; Zhai, X.; Unterthiner, T.; Dehghani, M.; Minderer, M.; Heigold, G.; Gelly, S.; Uszkoreit, J.; and Houlsby, N. 2020.
\newblock An {Image} is {Worth} 16x16 {Words}: {Transformers} for {Image} {Recognition} at {Scale}.
\newblock ArXiv: 2010.11929.

\bibitem[{Dresdner et~al.(2023)Dresdner, Kochkov, Norgaard, Zepeda-Núñez, Smith, Brenner, and Hoyer}]{dresdner_learning_2023}
Dresdner, G.; Kochkov, D.; Norgaard, P.; Zepeda-Núñez, L.; Smith, J.~A.; Brenner, M.~P.; and Hoyer, S. 2023.
\newblock Learning to correct spectral methods for simulating turbulent flows.
\newblock ArXiv:2207.00556 [cs].

\bibitem[{Duruisseaux et~al.(2024)Duruisseaux, Liu-Schiaffini, Berner, and Anandkumar}]{duruisseaux_towards_2024}
Duruisseaux, V.; Liu-Schiaffini, M.; Berner, J.; and Anandkumar, A. 2024.
\newblock Towards {Enforcing} {Hard} {Physics} {Constraints} in {Operator} {Learning} {Frameworks}.
\newblock In \emph{ICML 2024 AI for Science Workshop}.

\bibitem[{Gao, Kaltenbach, and Koumoutsakos(2025)}]{gao2025generative}
Gao, H.; Kaltenbach, S.; and Koumoutsakos, P. 2025.
\newblock Generative learning of the solution of parametric partial differential equations using guided diffusion models and virtual observations.
\newblock \emph{Computer Methods in Applied Mechanics and Engineering}, 435: 117654.

\bibitem[{Geneva and Zabaras(2020)}]{geneva2020modeling}
Geneva, N.; and Zabaras, N. 2020.
\newblock Modeling the dynamics of PDE systems with physics-constrained deep auto-regressive networks.
\newblock \emph{Journal of Computational Physics}, 403: 109056.

\bibitem[{Guo, Li, and Iorio(2016)}]{guo2016convolutional}
Guo, X.; Li, W.; and Iorio, F. 2016.
\newblock Convolutional neural networks for steady flow approximation.
\newblock In \emph{Proceedings of the 22nd ACM SIGKDD international conference on knowledge discovery and data mining}, 481--490.

\bibitem[{Hao et~al.(2024)Hao, Su, Liu, Berner, Ying, Su, Anandkumar, Song, and Zhu}]{hao2024dpot}
Hao, Z.; Su, C.; Liu, S.; Berner, J.; Ying, C.; Su, H.; Anandkumar, A.; Song, J.; and Zhu, J. 2024.
\newblock Dpot: Auto-regressive denoising operator transformer for large-scale pde pre-training.
\newblock \emph{arXiv preprint arXiv:2403.03542}.

\bibitem[{Hao et~al.(2023)Hao, Wang, Su, Ying, Dong, Liu, Cheng, Song, and Zhu}]{hao_gnot_2023}
Hao, Z.; Wang, Z.; Su, H.; Ying, C.; Dong, Y.; Liu, S.; Cheng, Z.; Song, J.; and Zhu, J. 2023.
\newblock {GNOT}: {A} {General} {Neural} {Operator} {Transformer} for {Operator} {Learning}.
\newblock In \emph{Proceedings of the 40th {International} {Conference} on {Machine} {Learning}}, 12556--12569. PMLR.
\newblock ISSN: 2640-3498.

\bibitem[{Hendrycks and Gimpel(2016)}]{hendrycks_gaussian_2016}
Hendrycks, D.; and Gimpel, K. 2016.
\newblock Gaussian {Error} {Linear} {Units} ({GELUs}).
\newblock ArXiv:1606.08415 [cs].

\bibitem[{Hu et~al.(2024)Hu, Wang, Zheng, Zhang, Feng, Feng, Wei, Wang, Ma, and Wu}]{hu_wavelet_2024}
Hu, P.; Wang, R.; Zheng, X.; Zhang, T.; Feng, H.; Feng, R.; Wei, L.; Wang, Y.; Ma, Z.-M.; and Wu, T. 2024.
\newblock Wavelet {Diffusion} {Neural} {Operator}.
\newblock ArXiv:2412.04833 [cs].

\bibitem[{Huang et~al.(2024)Huang, Shi, Gao, Wei, Zhang, Bian, Yang, and Liu}]{huang2024lordnet}
Huang, X.; Shi, W.; Gao, X.; Wei, X.; Zhang, J.; Bian, J.; Yang, M.; and Liu, T.-Y. 2024.
\newblock LordNet: An efficient neural network for learning to solve parametric partial differential equations without simulated data.
\newblock \emph{Neural Networks}, 176: 106354.

\bibitem[{Huang et~al.(2023)Huang, Shi, Meng, Wang, Gao, Zhang, and Liu}]{huang_neuralstagger_2023}
Huang, X.; Shi, W.; Meng, Q.; Wang, Y.; Gao, X.; Zhang, J.; and Liu, T.-Y. 2023.
\newblock {NeuralStagger}: {Accelerating} {Physics}-constrained {Neural} {PDE} {Solver} with {Spatial}-temporal {Decomposition}.
\newblock In \emph{Proceedings of the 40th {International} {Conference} on {Machine} {Learning}}, 13993--14006. PMLR.
\newblock ISSN: 2640-3498.

\bibitem[{Jha(2024)}]{jha_residual-based_2024}
Jha, P.~K. 2024.
\newblock Residual-{Based} {Error} {Corrector} {Operator} to {Enhance} {Accuracy} and {Reliability} of {Neural} {Operator} {Surrogates} of {Nonlinear} {Variational} {Boundary}-{Value} {Problems}.
\newblock \emph{Computer Methods in Applied Mechanics and Engineering}, 419: 116595.
\newblock ArXiv:2306.12047 [math].

\bibitem[{Jha and Oden(2022)}]{jha_goal-oriented_2022}
Jha, P.~K.; and Oden, J.~T. 2022.
\newblock Goal-oriented a-posteriori estimation of model error as an aid to parameter estimation.
\newblock \emph{Journal of Computational Physics}, 470: 111575.

\bibitem[{Jiang et~al.(2020)Jiang, Kashinath, Prabhat, and Marcus}]{jiang_enforcing_2020}
Jiang, C.~M.; Kashinath, K.; Prabhat; and Marcus, P. 2020.
\newblock Enforcing {Physical} {Constraints} in {CNNs} through {Differentiable} {PDE} {Layer}.
\newblock In \emph{ICLR 2020 Workshop on Integration of Deep Neural Models and Differential Equations}.

\bibitem[{Li et~al.(2020{\natexlab{a}})Li, Kovachki, Azizzadenesheli, Liu, Bhattacharya, Stuart, and Anandkumar}]{li_fourier_2020}
Li, Z.; Kovachki, N.; Azizzadenesheli, K.; Liu, B.; Bhattacharya, K.; Stuart, A.; and Anandkumar, A. 2020{\natexlab{a}}.
\newblock Fourier {Neural} {Operator} for {Parametric} {Partial} {Differential} {Equations}.
\newblock ArXiv: 2010.08895.

\bibitem[{Li et~al.(2020{\natexlab{b}})Li, Kovachki, Azizzadenesheli, Liu, Bhattacharya, Stuart, and Anandkumar}]{li_neural_2020}
Li, Z.; Kovachki, N.; Azizzadenesheli, K.; Liu, B.; Bhattacharya, K.; Stuart, A.; and Anandkumar, A. 2020{\natexlab{b}}.
\newblock Neural {Operator}: {Graph} {Kernel} {Network} for {Partial} {Differential} {Equations}.
\newblock Number: arXiv:2003.03485 arXiv:2003.03485 [cs, math, stat].

\bibitem[{Lippe et~al.(2023)Lippe, Veeling, Perdikaris, Turner, and Brandstetter}]{lippe_pde-refiner_2023}
Lippe, P.; Veeling, B.~S.; Perdikaris, P.; Turner, R.~E.; and Brandstetter, J. 2023.
\newblock {PDE}-{Refiner}: {Achieving} {Accurate} {Long} {Rollouts} with {Neural} {PDE} {Solvers}.
\newblock ArXiv:2308.05732 [cs].

\bibitem[{Patera(1984)}]{patera1984spectral}
Patera, A.~T. 1984.
\newblock A spectral element method for fluid dynamics: laminar flow in a channel expansion.
\newblock \emph{Journal of computational Physics}, 54(3): 468--488.

\bibitem[{Reddy(1993)}]{reddy1993introduction}
Reddy, J.~N. 1993.
\newblock An introduction to the finite element method.
\newblock \emph{New York}, 27(14).

\bibitem[{Rosofsky, Majed, and Huerta(2022)}]{rosofsky2022applications}
Rosofsky, S.~G.; Majed, H.~A.; and Huerta, E. 2022.
\newblock Applications of physics informed neural operators.
\newblock \emph{arXiv preprint arXiv:2203.12634}.

\bibitem[{Sanchez-Gonzalez et~al.(2020)Sanchez-Gonzalez, Godwin, Pfaff, Ying, Leskovec, and Battaglia}]{Sanchez-Gonzalez2020}
Sanchez-Gonzalez, A.; Godwin, J.; Pfaff, T.; Ying, R.; Leskovec, J.; and Battaglia, P.~W. 2020.
\newblock Learning to {Simulate} {Complex} {Physics} with {Graph} {Networks}.
\newblock ArXiv: 2002.09405.

\bibitem[{Smith(1985)}]{smith1985numerical}
Smith, G.~D. 1985.
\newblock \emph{Numerical solution of partial differential equations: finite difference methods}.
\newblock Oxford university press.

\bibitem[{Wandel, Weinmann, and Klein(2020)}]{Wandel2020}
Wandel, N.; Weinmann, M.; and Klein, R. 2020.
\newblock Learning {Incompressible} {Fluid} {Dynamics} from {Scratch} -- {Towards} {Fast}, {Differentiable} {Fluid} {Models} that {Generalize}.
\newblock ArXiv: 2006.08762.

\bibitem[{Wandel, Weinmann, and Klein(2021)}]{wandel2021teaching}
Wandel, N.; Weinmann, M.; and Klein, R. 2021.
\newblock Teaching the incompressible Navier--Stokes equations to fast neural surrogate models in three dimensions.
\newblock \emph{Physics of Fluids}, 33(4).

\bibitem[{Wang et~al.(2025)Wang, Seidman, Sankaran, Wang, Pappas, and Perdikaris}]{wang_cvit_2025}
Wang, S.; Seidman, J.~H.; Sankaran, S.; Wang, H.; Pappas, G.~J.; and Perdikaris, P. 2025.
\newblock {CViT}: {Continuous} {Vision} {Transformer} for {Operator} {Learning}.
\newblock ArXiv:2405.13998 [cs].

\bibitem[{Wu et~al.(2024)Wu, Luo, Wang, Wang, and Long}]{wu_transolver_2024}
Wu, H.; Luo, H.; Wang, H.; Wang, J.; and Long, M. 2024.
\newblock Transolver: {A} {Fast} {Transformer} {Solver} for {PDEs} on {General} {Geometries}.
\newblock ArXiv:2402.02366.

\bibitem[{Wu and He(2018)}]{wu2018group}
Wu, Y.; and He, K. 2018.
\newblock Group normalization.
\newblock In \emph{Proceedings of the European conference on computer vision (ECCV)}, 3--19.

\bibitem[{Zhou et~al.(2024)Zhou, Ma, Wu, Wang, and Long}]{zhou2024unisolver}
Zhou, H.; Ma, Y.; Wu, H.; Wang, H.; and Long, M. 2024.
\newblock Unisolver: PDE-conditional transformers are universal PDE solvers.
\newblock \emph{arXiv preprint arXiv:2405.17527}.

\bibitem[{Zhu et~al.(2019)Zhu, Zabaras, Koutsourelakis, and Perdikaris}]{zhu2019physics}
Zhu, Y.; Zabaras, N.; Koutsourelakis, P.-S.; and Perdikaris, P. 2019.
\newblock Physics-constrained deep learning for high-dimensional surrogate modeling and uncertainty quantification without labeled data.
\newblock \emph{Journal of Computational Physics}, 394: 56--81.

\end{thebibliography}

\clearpage
\appendix
\onecolumn
\section{PDE Residuals Approximation}
\label{app:pde_loss}

Regarding a general PDE in the form (Equation~\ref{equ:governpde}), it can be further decomposed as
\begin{equation}
\label{equ:speratepde}
\mathcal{M}(\mathbf{u}_t, \mathbf{u}_{tt})=\mathcal{L}(\mathbf{u}) + \mathcal{N}(\mathbf{u}),
\end{equation}
where $\mathcal{L}$ denotes the linear term (e.g., diffusion, wave operator) and $\mathcal{N}$ denotes the non-linear term (e.g., advection). 
To approximate $\frac{\partial L_{\text{PDE}}}{\partial \hat{\mathbf{u}}_{t+1}}$ and enable the proposed caching strategy for efficiency, we generally treat linear terms in the PDE implicitly to ensure stability, while handling nonlinear terms explicitly to keep the Jacobian constant across timesteps. 
Under this semi-implicit scheme, the approximate Jacobian matrix becomes $\frac{I}{\Delta t}-\mathcal{L}$ for first-order equations or $\frac{I}{\Delta t^2}-\mathcal{L}$ for second-order equations.

We detail the numerical discretization employed in this work for defining the PDE residuals, along with relevant implementation considerations. 
In all experiments, the computational domain $\Omega\subseteq\mathbb{R}^2$ is discretized into an $n\times n$ uniform Cartesian grid with mesh spacing $\Delta$.
Any function defined on $\Omega$ is then approximated by its values at the grid points.
Specifically, we denote the discrete approximation of a function $u(x,y)$ at the grid point $(x_i, y_j)$ by $u_{i,j}$, 
where $x_i=x_0+i\Delta$ and $y_j=y_0+j\Delta$ for $i=0,\cdots,n-1$ and $j=0,\cdots,n-1$.
For time-dependent variables, we discretize the time interval $[0, T]$ with a uniform time step $\Delta t$, and denote a discrete function $u$ at the spatial grid point $(i,j)$ and $t$-th time step by $u_{i,j}^t$.

\paragraph{2D Navier-Stokes equation.}  Considering incompressible two-dimensional Navier-Stokes equations with periodic boundary conditions as follows: 
\begin{equation}
    \label{equ:nse}
\begin{gathered}
\frac{\partial \omega}{\partial t}=-\frac{\partial \psi}{\partial y} \frac{\partial \omega}{\partial x}+\frac{\partial \psi}{\partial x} \frac{\partial \omega}{\partial y}+\frac{1}{\operatorname{Re}}\left(\frac{\partial^2 \omega}{\partial x^2}+\frac{\partial^2 \omega}{\partial y^2}\right) + f(x,y),\\
\frac{\partial^2 \psi}{\partial x^2}+\frac{\partial^2 \psi}{\partial y^2}=-\omega,
\end{gathered}
\end{equation}
in which both vorticity function $\omega$ and steam function $\psi$ are time-dependent, and $f$ is the forcing term. 
Given the stream function $\psi^t_{i,j}$ at time step $t$ and a one-step prediction $\hat{\psi}^{t+1}_{i,j}$ for $i=1,\cdots,n$ and $j=1,\cdots,n$,
we first apply central difference based differential operator to compute the corresponding vorticities $\omega^t_{i,j}$ and $\hat{\omega}^{t+1}_{i,j}$.
To integrate the governing equation (Equation~\ref{equ:nse}), we adopt an implicit scheme for the diffusion term to ensure numerical stability, while an explicit scheme for the advection part to avoid non-linearity. 
Using a forward Euler method in time, the PDE residual for Equation~\ref{equ:nse} used to approximate the Jacobian matrix is then defined as follows,
\begin{equation}
\begin{aligned}
L_{\text{PDE}} = 
\sum_{i,j}^{n\times n} 
  \Big|\frac{1}{\Delta t}(-\hat{\omega}^{t+1}_{i,j} + \omega^t_{i,j}) 
  - \frac{1}{4\Delta^2} \left[
    \left(\psi_{i,j+1}^t - \psi_{i,j-1}^t\right) \right.
    \left. \left(\omega_{i+1,j}^t - \omega_{i-1,j}^t\right) \right.\\
    \left.
    - \left(\psi_{i+1,j}^t - \psi_{i-1,j}^t\right)
    \left(\omega_{i,j+1}^t - \omega_{i,j-1}^t\right)
  \right] \\
  + 0.5 \frac{1}{\text{Re} \Delta^2} \left(
    \omega_{i+1,j}^t + \omega_{i-1,j}^t + \omega_{i,j+1}^t + \omega_{i,j-1}^t - 4\omega_{i,j}^t \right. \\
    \left.
    + \hat{\omega}_{i+1,j}^{t+1} + \hat{\omega}_{i-1,j}^{t+1} + \hat{\omega}_{i,j+1}^{t+1} + \hat{\omega}_{i,j-1}^{t+1} - 4\hat{\omega}_{i,j}^{t+1}
  \right) + f_{i,j}\Big|
\end{aligned}
\end{equation}
We can see that $\frac{\partial L_{\text{PDE}}}{\partial \hat{\psi}^{t+1}}$ depends only on the current state $\psi^t$.

\paragraph{2D Wave Equation.} The two-dimensional wave equation with constant wave speed is formulated as
\begin{equation}
    \label{equ:wave}
    \frac{\partial^2 u}{\partial t^2}=c^2(\frac{\partial^2 u}{\partial x^2} + \frac{\partial^2 u}{\partial y^2}),
\end{equation}
where the displacement $u$ is time-dependent and $c$ is the constant wave speed. 
We use the second-order central difference to approximate the Laplace operator here, so we have
\begin{equation}\label{eq:discrete_poisson}
    \begin{aligned}
           \frac{\partial^2u}{\partial t^2}\Big|_{i,j}= c^2\left(\frac{\partial^2 u}{\partial x^2} + \frac{\partial^2 u}{\partial y^2}\right)_{i,j} = &
            \frac{u_{i-1,j} - 2u_{i,j} + u_{i+1,j}}{\Delta^2} + \frac{u_{i,j-1} - 2u_{i,j} + u_{i,j+1}}{\Delta^2} ,
    \end{aligned}
\end{equation}
at each grid point $(x_i,y_j)$ for $i=1,\cdots,n$ and $j=1,\cdots,n$.
Since we did not have fine sampling in the temporal direction, we use an implicit scheme and central finite difference for the second-order time derivative.
Given the displacement function $u^{t-1}_{i,j}$ at time $t-1$ and $u^{t}_{i,j}$ at time $t$, and the one-step prediction $\hat{u}^{t+1}_{i,j}$ at time $t+1$,  the discretized PDE residual is represented as:
\begin{align}
    \label{equ:wave_pde}
    L_{\text{PDE}}=\sum_{i,j}^{n\times n}\Big|
     & \frac{\hat{u}^{t+1}_{i,j} + u^{t-1}_{i,j} - 2u^t_{i,j}}{\Delta t^2}- \notag \\
     & \frac{c^2}{3}\left(\frac{\partial^2 u^{t-1}}{\partial x^2} + \frac{\partial^2 u^{t-1}}{\partial y^2} + \frac{\partial^2 u^{t}}{\partial x^2} + \frac{\partial^2 u^{t}}{\partial y^2} + \frac{\partial^2 \hat{u}^{t+1}}{\partial x^2} + \frac{\partial^2 \hat{u}^{t+1}}{\partial y^2}\right)_{i,j}\Big|.
\end{align}
Hence, $\frac{\partial L_{\text{PDE}}}{\partial \hat{u}^{t+1}}$ does not depend on the prediciton $\hat{u}^{t+1}$ itself. Notably, for the wave equation, being a linear system, the PDE residual in Equation~\ref{equ:taylor} does not rely on the assumption of small prediction error, as no Taylor expansion is needed. This further implies that PhysicsCorrect remains robust to prediction errors in linear systems, a phenomenon that will be demonstrated empirically in later sections.

\paragraph{Kuramoto-Sivashinsky Equation.} The Kuramoto-Sivashinsky equation with a constant viscosity parameter of 1.0 is as follows: 
\begin{equation}
    \label{equ:ks}
     v_t+v_{x x}+v_{x x x x}+v v_x=0.
\end{equation}
We apply the spectral method to obtain the spatial derivative for different orders.
Given $v^t$ at $t$-th time step and a one-step prediction $\hat{v}^{t+1}$ at $t+1$-th time step, the PDE residual is defined as follows:
\begin{equation}
    \label{equ:pde_ks}
    L_{\text{PDE}}=\sum_{i}^{n} \Big|\frac{\hat{v}^{t+1}_i - v^t_i}{\Delta t}+0.5(\frac{\partial^2 v^t}{\partial x^2}+\frac{\partial^4 v^t}{\partial x^4}+v^t \frac{\partial v^t}{\partial x} +\frac{\partial^2 \hat{v}^{t+1}}{\partial x^2}+\frac{\partial^4 \hat{v}^{t+1}}{\partial x^4}+\hat{v}^{t+1} \frac{\partial \hat{v}^{t+1}}{\partial x})_{i}\Big|,
\end{equation}
where $i$ denotes the spatial points. 
We simplify this by pre-computing the pseudoinverse of the Jacobian matrix using a semi-implicit discretization for the nonlinear term and an implicit scheme for the linear term, which approximates the time-dependent Jacobian $\frac{\partial L_{\text{PDE}}}{\partial \hat{v}^{t+1}}$. 
In the subsequent correction steps, all terms in the PDE residual are treated implicitly to further improve the accuracy of the residual formulation.

\begin{figure}[!htb]
    \centering
    \includegraphics[width=0.6\linewidth]{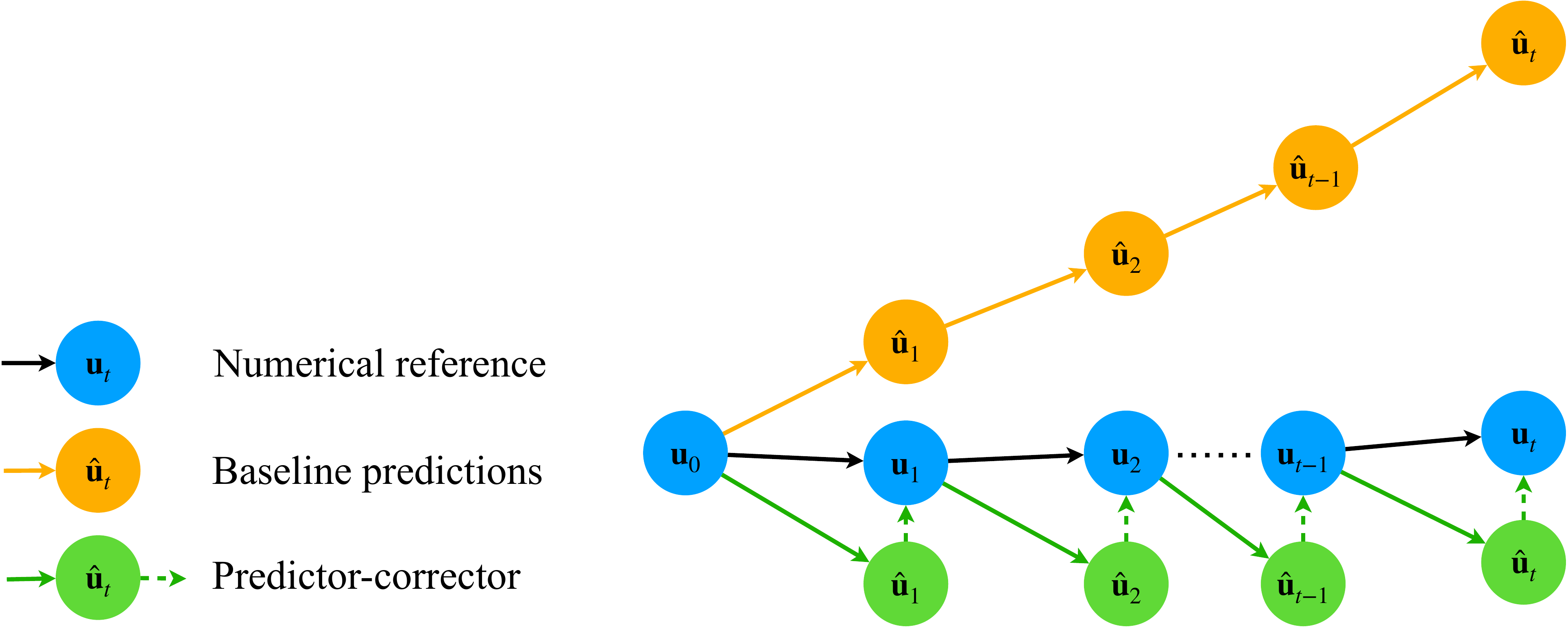}
    \caption{Schematic illustration of the PhysicsCorrect predictor–corrector workflow. Starting from the true trajectory $(\mathbf{u}_0, \dots, \mathbf{u}_t)$, a pretrained neural PDE solver produces baseline predictions $(\hat{\mathbf{u}}_1, \dots, \hat{\mathbf{u}}_t)$, which gradually deviate from the numerical reference. Our method applies a physics-informed correction at each rollout step, projecting $\hat{\mathbf{u}}_t$ back toward the PDE-consistent manifold and preventing long-term error accumulation.}
    \label{fig:schematic_diagram}
\end{figure}

\paragraph{Heuristic Theoretical Analysis.}
We provide a heuristic theoretical analysis to offer intuitive insight into the behavior of the proposed method (A schematic diagram is shown in Figure~\ref{fig:schematic_diagram}). 
Under the linearized correction, let $r = L_{\text{PDE}}(\mathbf{u}_t, \hat{\mathbf{u}}_{t+1})$ denote the current PDE residual, and $r^{+} = L_{\text{PDE}}(\mathbf{u}_t, \hat{\mathbf{u}}_{t+1} + \mathbf{u}_c)$ the updated residual after applying the correction $\mathbf{u}_c$.
Linearizing around $\hat{\mathbf{u}}_{t+1}$ gives
\begin{equation}
r^{+} \approx r + A \mathbf{u}_c, \qquad
A = \frac{\partial L_{\text{PDE}}}{\partial \hat{\mathbf{u}}_{t+1}}.
\end{equation}
Solving it in the least-squares sense yields
\begin{equation}
    \mathbf{u}_c^* = -A^{\dagger}r, \quad
r^{+} \approx (I - A A^{\dagger})r,
\end{equation}
where $A^{\dagger}$ is the Moore--Penrose pseudoinverse and $A A^{\dagger}$ the orthogonal projector onto $\mathrm{range}(A)$.
Each correction step therefore removes the component of the residual lying in $\mathrm{range}(A)$, strictly reducing its norm unless $r \perp \mathrm{range}(A)$.
This shows that the correction projects predictions back onto the PDE-consistent manifold, mitigating error accumulation in long rollouts. With a relaxed update $\mathbf{u}_c = -\gamma A^{\dagger}r$ ($0 < \gamma < 2$), under which
\[
r^{+} = (I - \gamma A A^{\dagger})r
\]
acts as a contraction on $\mathrm{range}(A)$, ensuring residual decay and connecting the theoretical bound to our empirical stability observations.

\section{Experimental Details} 
\label{app:exp_detail}
In this section, we introduce the details of the configurations of different benchmarks and neural network configurations, as well as additional visualization results. 

\subsection{Navier-Stokes Equation}
\label{app:exp_detail:nse}
We generate 1,000 simulation trajectories as training data using a Reynolds number of 1,000 and a forcing term defined as $f(x,y)=0.1\sin(2\pi(x+y))+\cos(2\pi(x+y))$. 
Each trajectory begins from a Gaussian random vorticity field sampled from the distribution $\mathcal{N}\left(0,8^{3}(-\Delta+64 I)^{-4.0}\right)$. 
An additional 64 trajectories, generated using the same configuration, are used for testing.
For training, we include the first 100 time steps of each trajectory, while evaluation measures generalization performance over 1,000 time steps. The neural network is trained to predict the one-step residual $\delta \psi^t=\psi^{t+1} - \psi^{t}$. 
We use an initial learning rate of 1e-3 and a weight decay of 1e-5 for the Adam optimizer, and decay the learning rate by a factor of 0.9 every 5,000 iterations.
 
Regarding the neural network architectures, we implement the standard Fourier Neural Operator (FNO) following \citep{li_neural_2020}, which consists of four Fourier layers. 
Each layer uses 12 Fourier modes in both the $x$ and $z$ directions, and the width of the feature map is set to 64. 
The activation function used is the Gaussian Error Linear Unit (GELU) \citep{hendrycks_gaussian_2016}. 
After the Fourier layers, a linear projection is applied to map the high-dimensional latent representation back to the spatial domain.
For the UNet architecture, we adopt the configuration described in \citep{huang_neuralstagger_2023}, with a hidden size of 128.
For the Vision Transformer (ViT), we use a patch size of 8 for spatial embedding and stack 4 standard pre-norm transformer blocks. 
Each block contains four attention heads, with each head having a dimensionality of 128. 
GELU is also used as the activation function for ViT.

\begin{figure}[!htb]
    \centering
    \includegraphics[width=1.0\linewidth]{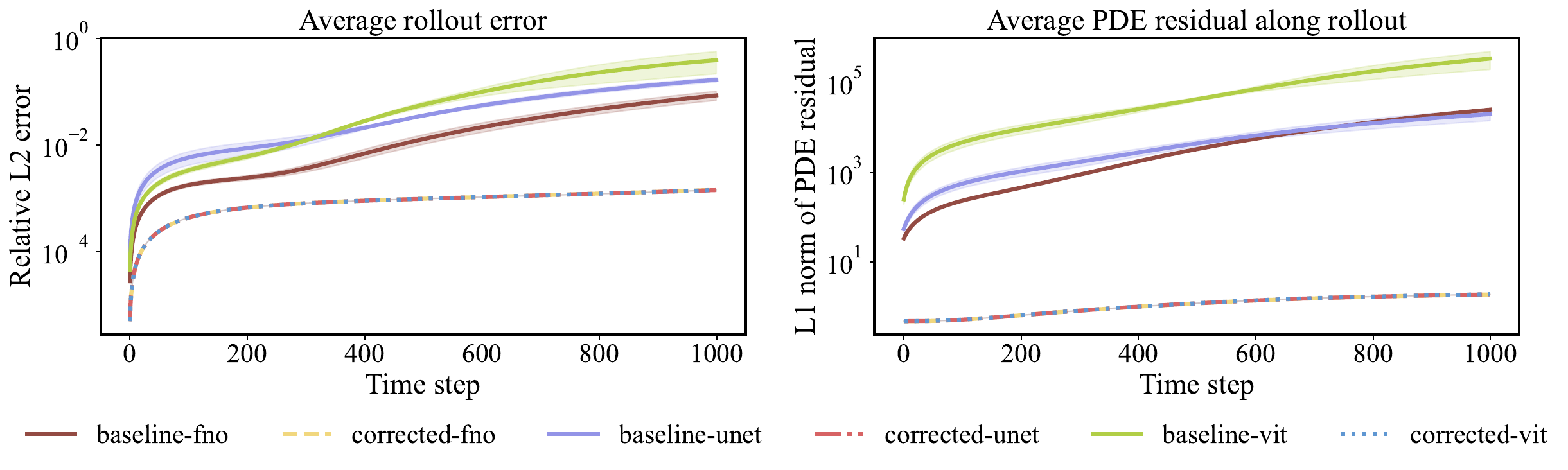}
    \caption{Long-term rollout performance on the 2D Navier-Stokes equation across different neural architectures. Left: Relative L2 error over 1,000 time steps, showing that baseline models (solid lines) suffer from error accumulation and eventual divergence, while our corrected models (dashed lines) maintain stable and accurate predictions throughout the simulation. Right: PDE residual magnitude, where lower values indicate better satisfaction of the governing equation. Results are averaged over 5 random seeds, with shaded regions showing standard deviation.}
    \label{fig:nse-perf-comp}
\end{figure}

After training for 30,000 iterations with a batch size of 128, we evaluate different baseline models and their corresponding corrected simulation trajectories.
Figure~\ref{fig:nse-perf-comp} presents the relative L2 norm error (computed by dividing the error norm by the ground-truth norm) and the PDE residual histories over the rollout horizon. While all baseline models suffer from error accumulation over time, the proposed corrector yields stable and accurate long-term predictions. The residual histories further confirm that our method consistently enforces physical constraints at each time step.

\subsection{2D Wave Equation}
\label{app:exp_detail:wave}
We fixed the wave speed $c$ as 1.0 in our test. 
We generate Gaussian random fields as initial conditions and then use fourth-order central differences for spatial derivatives and a fourth-order Runge–Kutta (RK4) scheme for time integration, to accurately evolve the displacement field, following \citep{rosofsky2022applications}.
The data are generated on a 128$\times$128 grid with a time step of 1e-4 and recorded every 100 time steps.
We sampled 512 trajectories with the first 10 steps (with an interval of 1e-2) for training and another 64 trajectories with all 100 steps for testing. 
We use an initial learning rate of 1e-3 with a decay factor of 0.6 every 100 epochs.

The FNO configuration for the wave equation is identical to that used for the Navier–Stokes equation. 
However, the UNet used in this case has a reduced hidden size of 20 and uses LeakyReLU activation function with a negative slope of 0.1.
For the ViT, we use a patch size of 4 and increase the embedding dimension to 256. 
The model consists of 3 transformer blocks, each with eight attention heads and a head dimension of 256.

\begin{figure}[!htb]
    \centering
    \includegraphics[width=1.\linewidth]{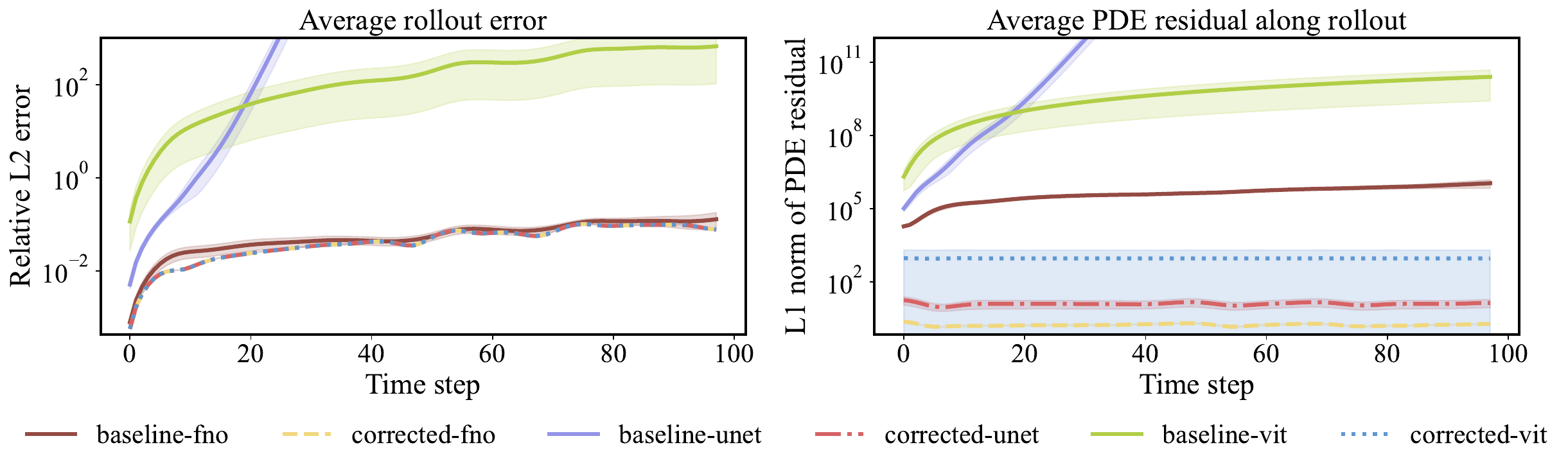}
    \caption{Long-term rollout performance on the 2D wave equation using second-order residual prediction. Left: Relative L2 error over 100 time steps for different neural architectures with (dashed lines) and without (solid lines) our physics-informed corrector. Right: PDE residual magnitude during rollout. While second-order formulation already provides reasonable baseline performance (especially for FNO), our correction framework consistently reduces both prediction error and PDE residual across all architectures. Results are averaged over 5 random seeds with shaded regions showing standard deviation.}
    \label{fig:wave-sample-various-baseline-2nd}
\end{figure}

Next, we train different baseline models to predict the second-order residual for 1,000 epochs with a batch size of 128.
Figure~\ref{fig:wave-sample-various-baseline-2nd} shows the relative L2 error and PDE residual histories over 100 time steps.
While the improvement of our approach on the FNO baseline is modest, significant gains are observed for the UNet and ViT baselines.
Once again, the PDE residual plots indicate that our approach consistently enforces physical fidelity at each time step.

\begin{figure}
    \centering
    \includegraphics[width=1.\linewidth]{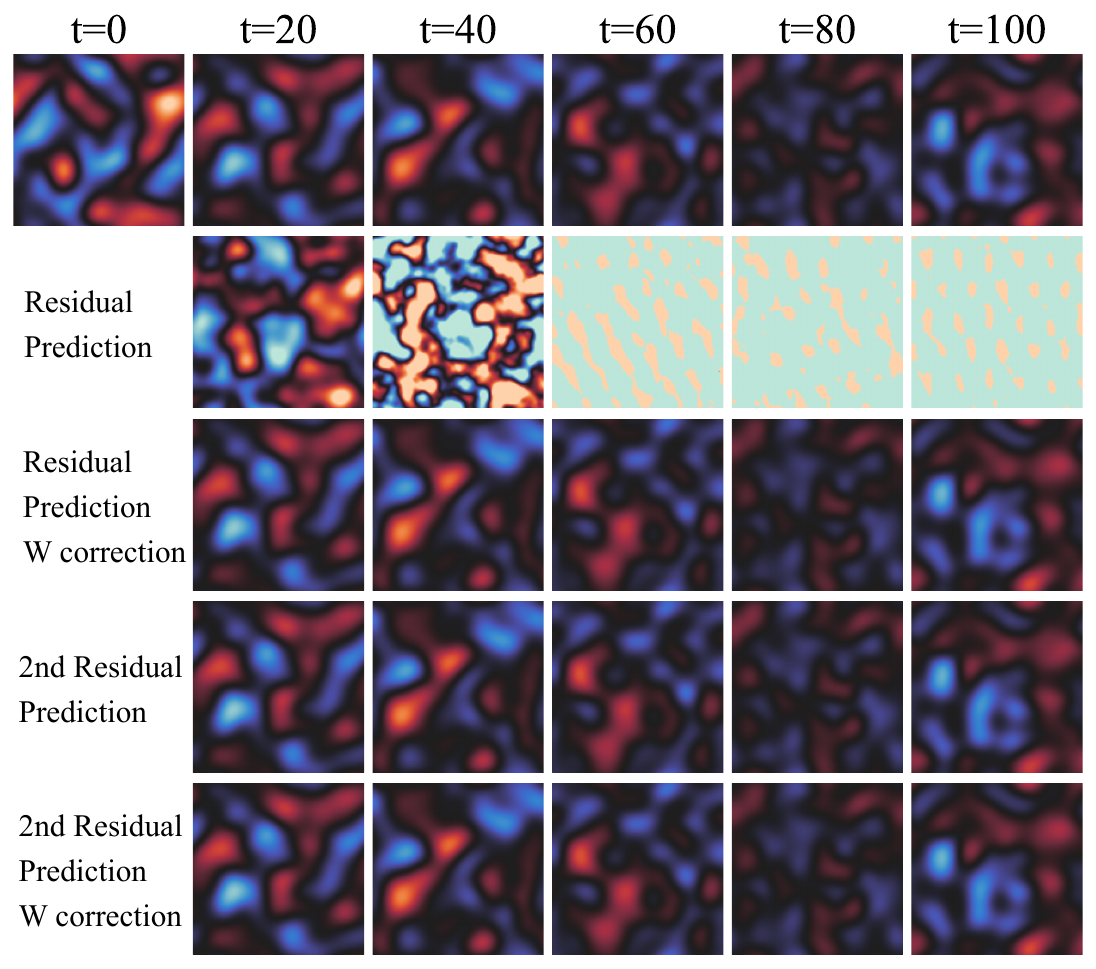}
    \caption{Comparison of rollout predictions for a representative test sample using different network output strategies and correction methods. The top row shows the ground-truth evolution from $t=0$ to $t=100$. The second row shows the result of first-order residual prediction (trained with full horizon), which rapidly diverges. The third row applies our proposed correction to the first-order residual prediction, significantly improving stability. The fourth and fifth rows show the predictions using second-order residuals (trained with the first 10 time steps), with and without correction, both yielding accurate and stable long-term rollouts.}
    \label{fig:wave_fno_different_output}
\end{figure}

\paragraph{First-order residual vs second-order residual.}
As mentioned earlier, the choice between predicting the first-order residual (predicting $u^{t+1}-u^{t}$) or the second-order residual (predicting $\delta u^t=u^{t+1} + u^{t-1} - 2u^t$) results in noticeably different long-rollout behaviors, significantly affecting rollout stability and error accumulation.
Here, we visualize the result of a representative test sample using FNO, as shown in Figure~\ref{fig:wave_fno_different_output}.
The baseline rollout with first-order residual prediction diverges rapidly, whereas the rollout with the proposed corrector remains stable.
When the network is trained to predict the second-order residual, both the baseline model and the corrected version exhibit stable performance.
However, the improvements in this case are marginal for two reasons:
(1) the baseline model already performs well and closely satisfies the PDE residual; and (2) discretization-induced errors, as discussed in Figure~\ref{fig:nse_ref_pde}, impose an inherent limitation on correction accuracy.

\subsection{Kuranmoto-Sivashinksy Equation}
\label{app:exp_detail:ks}
We aim to solve the Equation~\ref{equ:ks} on a temporal domain $[0, 100]$ and a spatial domain $[0,64]$ while focusing on the chaotic part. 
The equation is numerically solved using the spectral method \citep{dresdner_learning_2023}, with a spatial interval of 0.125 and a temporal interval of 0.05, for data generation.
We begin from $v(x,50)$ and train a one-step predictor using neural networks to learn the residual $\delta v^t=v^{t+1}-v^t$.
The training set includes 512 trajectories, using only the first 500 time steps of each, while the test set contains 64 additional trajectories, each with the full 1,000-step rollout.
All models are trained with a learning rate of 1e-3, a batch size of 256, and a total of 3,000 training epochs.

For the network configurations, we use a one-dimensional variant of the FNO (FNO-1D), adapted from its 2D counterpart by replacing all 2D convolutional operations with 1D convolutions. 
The number of Fourier modes is set to 16, and the feature map width is 64. 
GELU is used as the activation function.
The 1D UNet is similarly adapted from the 2D UNet by substituting all 2D convolutional layers with 1D versions. 
The hidden size is set to 64. 
We use the LeakyReLU activation function with a negative slope of 0.1, along with Group Normalization \citep{wu2018group}.
For the 1D ViT, we use a patch size of 4 and an embedding dimension of 768. 
The model consists of 4 transformer blocks, each containing 16 attention heads with a head dimension of 768. 
GELU is employed as the activation function throughout the model.

\begin{figure}[!htb]
    \centering
    \includegraphics[width=1.0\linewidth]{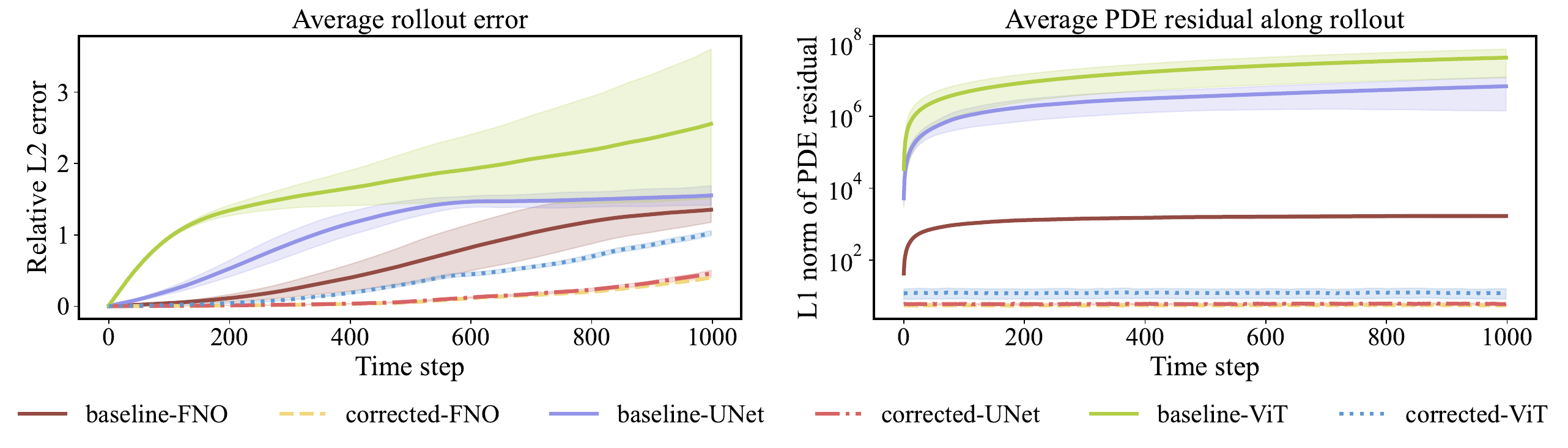}
    \caption{Long-term rollout performance on the KS equation across different neural architectures. Left: Relative L2 error over 1,000 time steps, showing that baseline models (solid lines) suffer from error accumulation and eventual divergence, while our corrected models (dashed lines) maintain relatively stable and accurate predictions throughout the simulation. Right: PDE residual magnitude, where lower values indicate better satisfaction of the governing equation. Results are averaged over 5 random seeds, with shaded regions showing standard deviation.}
    \label{fig:ks_various_baseline}
\end{figure}

For the corrector, we employ a semi-implicit scheme to precompute the pseudoinverse of the Jacobian matrix, and an implicit scheme for evaluating the PDE residual.
Figure~\ref{fig:ks_various_baseline} presents the relative L2 errors of the baseline models and their corrected counterparts, along with the corresponding PDE residual histories.
We observe that our approach consistently improves the rollout performance across all baseline models while maintaining physical consistency at each time step.
While applying an implicit scheme for both the pseudoinverse computation and PDE residual evaluation can further improve rollout accuracy (Figure~\ref{fig:ks_comparison}b), it incurs substantial computational overhead. Our approach strikes a balance between accuracy and efficiency.

While chaotic PDEs such as the KS equation exhibit extreme sensitivity to initial conditions, their trajectories remain confined to a low-dimensional manifold defined by the governing equations. Our correction step, even when based on an approximate and fixed Jacobian, serves as a coarse projection onto this manifold.
This projection regularizes the neural predictions by suppressing unstable high-frequency modes and redirects the state evolution toward dynamically consistent directions, preventing accumulation of off-manifold drift.
Therefore, perfect inversion of the true Jacobian is less critical than accurate residual definition.
As long as the correction step correlates with the true manifold-normal direction, stability can be maintained. This explains why our cached, approximate Jacobian yields substantial improvements even in chaotic regimes.

\subsection{Comparison with other baselines}
We compared PhysicsCorrect with PDE-Refiner \cite{lippe_pde-refiner_2023}. Under 1-D KS settings, PhysicsCorrect achieves over 3× lower cost, 3× faster inference, while reducing the 1000-step rollout error from 1.892 (PDE-refiner) to 0.461 (our approach), highlighting the efficiency of our method. PhysicsCorrect is orthogonal to training-based methods (e.g., DPOT \cite{hao2024dpot}) and can be seamlessly integrated with them.

\subsection{Additional Implementation details}

\paragraph{Compute Resources.} All training is performed on a single NVIDIA A6000 GPU. 
For the Navier-Stokes equation, training for 30,000 steps with a batch size of 64 takes roughly 60, 12, and 170 minutes using FNO, U-Net, and ViT, respectively.
For the wave equation, training for 1,000 epochs takes roughly 29, 6, and 65 minutes using FNO, U-Net, and ViT, respectively.
For the KS equation, training for 3,000 epochs takes roughly 100 minutes using FNO, 70 minutes using U-Net, and approximately 5 hours using ViT.

\paragraph{Negligible Additional Inference Time.} We also evaluate the inference time of both the baseline models and their counterparts augmented with our proposed correctors.
As shown in Figure~\ref{fig:results_summary}, the ratio between the inference time of the baseline with corrector and that of the baseline model is close to 1.0.
Although the Navier–Stokes case with the ViT baseline shows a slightly higher ratio, due to the lightweight nature of the baseline, the additional inference time introduced by our corrector remains negligible, particularly for larger-scale problems and more complex neural networks.

\subsection{Sensitivity Analysis: Error Structure and Magnitude Limits}
To characterize the operational bounds of PhysicsCorrect, we conducted a systematic sensitivity analysis by artificially corrupting ground truth solutions with controlled error patterns and magnitudes. This approach enables evaluation of corrector robustness under varying error conditions that may arise in practice.

\paragraph{Experimental Protocol.}
We simulate different types of prediction errors by adding two distinct noise patterns to reference solutions: uncorrelated Gaussian noise representing random, spatially independent perturbations, and correlated Gaussian random fields representing structured, spatially coherent perturbations, with controlled magnitude (relative L2 error ranging from $10^{-3}$ to 1.0). 
These corrupted states serve as surrogate neural network predictions, enabling systematic evaluation of corrector performance across different error characteristics. 

\begin{figure}[!htb]
    \centering
    \includegraphics[width=1.0\linewidth]{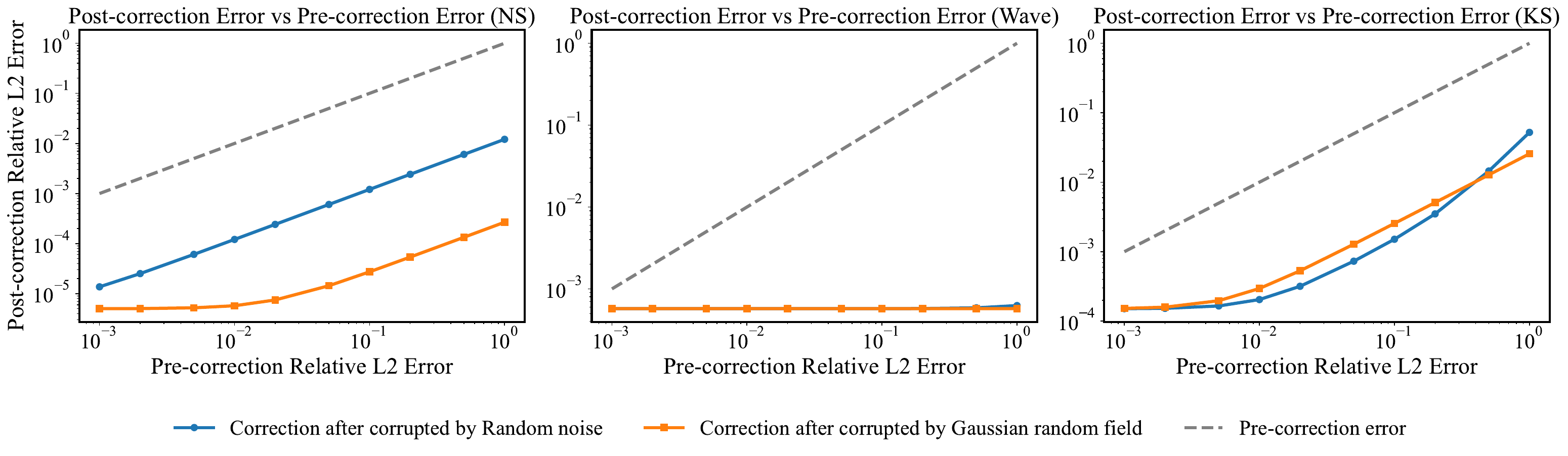}
    \caption{Post-correction relative L2 error versus pre-correction (one-step) relative L2 error across three PDE systems: Navier–Stokes (NS), Wave, and Kuramoto–Sivashinsky (KS). The x-axis represents the artificially injected corruption level before correction, while the y-axis shows the resulting error after applying the PhysicsCorrect method. Results are reported for both unstructured Gaussian noise and structured Gaussian random field perturbations. The dashed line indicates the uncorrected pre-correction error (i.e., the identity line), serving as a reference. Correction consistently reduces error magnitude, with stronger suppression observed for structured noise.}
    \label{fig:sensitivity_analysis}
\end{figure}

\paragraph{Linear versus Nonlinear PDE Response.}
Figure~\ref{fig:sensitivity_analysis} illustrates the post-correction relative L2 error achieved by the proposed PhysicsCorrect method versus the pre-correction error across different PDE systems.
The diagonal dashed line serves as a reference indicating no correction. 
Across all systems, our method significantly reduces the prediction error compared to the uncorrected baseline, particularly when the corruption is structured (Gaussian random field).
The analysis reveals a fundamental distinction between linear and nonlinear systems. 
For the wave equation, being a linear PDE with exact Jacobian computation, the corrector maintains effectiveness across the entire tested error range, indicating that the PhysicsCorrect module has likely reached the discretization error floor. 
The linear nature of the governing equation ensures that our correction mechanism remains theoretically sound regardless of error magnitude because the PDE residual is linear to the prediction, where Equation~\ref{equ:taylor} does not rely on the assumption of small relative error on the prediction for Taylor expansion.

In contrast, the nonlinear Navier-Stokes and Kuramoto-Sivashinsky equations exhibit threshold behavior. 
Within moderate error levels, correction accuracy remains high, demonstrating the robustness of PhysicsCorrect.
However, beyond certain thresholds, where the assumption of local approximation underlying the Taylor expansion no longer holds, the corrected relative error begins to increase with the pre-correction error.
This degradation is further attributed to the semi-implicit discretization schemes used for nonlinear PDEs, which yield approximate rather than exact Jacobians. As a result, the linearization becomes less valid under large perturbations.
Nevertheless, the results demonstrate that even when the corrected predictions are not back onto the manifold of physically consistent solutions, PhysicsCorrect still offers meaningful accuracy improvements.

\paragraph{Error Structure Dependencies.}
Correlated errors consistently prove more challenging for the corrector than uncorrelated noise, particularly in nonlinear chaotic systems (Rightmost panel of Figure~\ref{fig:sensitivity_analysis}). 
This observation is significant because neural network prediction errors typically exhibit spatial structure rather than random patterns, suggesting that real-world performance may be more constrained than random noise analysis would indicate.

\paragraph{Practical Implications.}
These findings establish that PhysicsCorrect's effectiveness is bounded by the validity of the underlying linearization approximation. 
For linear PDEs, this bound is theoretical, while for nonlinear systems, practical thresholds exist beyond which correction degrades. 
Users should therefore validate their neural operator's typical error characteristics against these bounds before deployment, and consider implementing error monitoring to detect when correction should be disabled since the original prediction fails.

\subsection{Sensitivity Analysis: Effect of Spatial Discretization and Resolution}
Beyond the sensitivity analysis of prediction accuracy to correction performance, as discussed in the main text, the accuracy of PDE residual evaluation also plays a key role in determining correction quality. Inaccurate residuals can degrade the overall performance of the prediction–correction pipeline. Both the spatial discretization scheme and the spatial resolution of the physical field strongly influence the accuracy of the residual calculation.
In this subsection, we further examine how discretization and resolution in the prediction step affect the performance of the prediction–correction framework.

\paragraph{Experimental Protocol.}
We use the wave equation with the FNO baseline as a benchmark, where the Jacobian matrix is accurately defined, allowing us to focus on the discretization and resolution aspects.
Following the same training settings as in the previous wave equation experiments, we evaluate our approach under different spatial discretization schemes: three‑point central finite difference (“standard”), five‑point central finite difference (“5‑point CFD”), and spectral method (“spectral”). We also assess performance at different spatial resolutions by down‑sampling the grid by factors of 2$\times$, 4$\times$, and 8$\times$.

\begin{figure}[!htb]
    \centering
    \includegraphics[width=1.0\linewidth]{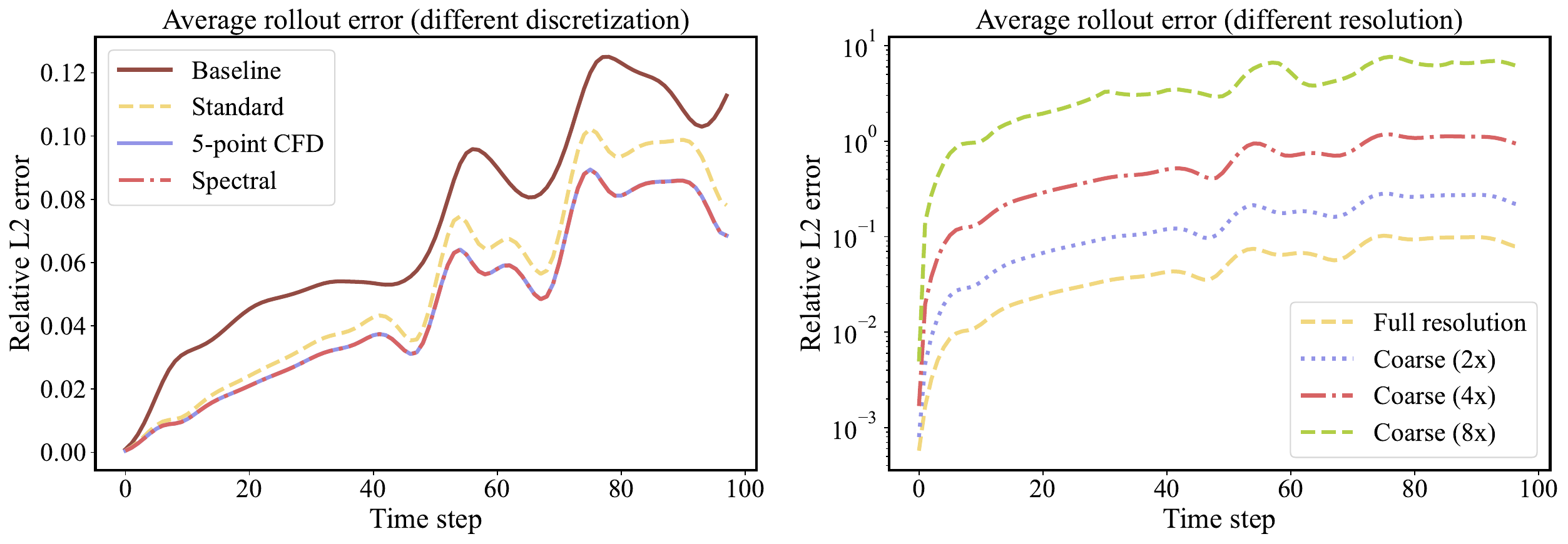}
    \caption{Sensitivity of prediction–correction error to spatial discretization schemes and spatial resolution. (Left) Average rollout error for different discretization methods. Higher‑accuracy discretization schemes lead to lower accumulated error than the standard three‑point central difference. (Right) Average rollout error for different spatial resolutions. Higher spatial resolution substantially reduces long‑term prediction error. Coarser resolutions (e.g., 4$\times$ and 8$\times$ down‑sampling) show a pronounced increase in error and faster error accumulation, though high‑accuracy discretization schemes retain some advantage even at coarse resolutions.}
    \label{fig:sensitivity_analysis_discretization_resolution}
\end{figure}

\paragraph{Effect of Discretization.}
As shown in Figure~\ref{fig:sensitivity_analysis_discretization_resolution} (left), higher‑accuracy discretization schemes consistently yield better prediction–correction performance, as each correction step produces a more accurate corrected state and keeps the predicted trajectory closer to the reference solution. While advanced discretization schemes appear similar at a glance, we observe small but non‑negligible differences, with average discrepancies below $3.5\times 10^{-5}$. These results indicate that improving the spatial discretization enhances one‑step correction accuracy, which in turn benefits long‑term stability.

\paragraph{Effect of Resolution.}
Similarly, Figure~\ref{fig:sensitivity_analysis_discretization_resolution} (right) shows that reducing the spatial resolution of both the prediction and the reference field makes PDE residual evaluation more challenging, leading to larger errors in subsequent correction steps. Coarser grids exacerbate error accumulation, ultimately degrading prediction–correction performance. The coarser the grid, the worse the accuracy. We note that this limitation could be mitigated by integrating spatio-temporal decomposition training frameworks \citep{huang_neuralstagger_2023}, preserving the large spatial- and temporal-interval advantages of purely neural PDE solvers while enhancing stability through physics-based correction. 

\paragraph{Practical Implications.} The results collectively demonstrate that both spatial discretization quality and spatial resolution are critical for achieving stable and accurate prediction–correction performance.  High‑accuracy discretization schemes improve the fidelity of PDE residual estimation, leading to more effective corrections and slower long‑term error growth. Likewise, higher spatial resolution mitigates residual evaluation errors and reduces the risk of error amplification during rollout. In practice, these two factors are intertwined: a coarse resolution can undermine the benefits of an advanced discretization scheme, whereas a high‑accuracy discretization can partly compensate for moderate resolution. Therefore, optimal performance is achieved when both discretization and resolution are chosen to balance computational cost with the accuracy requirements of the task.

\subsection{Potential scalability improvements}
In order to improve the scalability of the proposed approach, we conducted preliminary blockwise Jacobian–vector product experiments, which show a 34 \% reduction in memory and runtime with comparable accuracy at 128$^2$ resolution. 
The bottleneck stems from pseudoinverses rather than the correction principle. In addition, iterative CG solvers or structure-aware Jacobians would enable 3-D extension of our approach.

\section{Broader Impacts}
\label{sec:broaderimpact}
Time-dependent PDEs are fundamental to scientific and engineering applications including fluid dynamics, material modeling, seismic analysis, and climate forecasting. While neural PDE solvers offer computational efficiency, their long-term stability remains challenging for reliable predictions. Our physics-informed correction framework addresses this limitation without requiring model retraining, enabling seamless integration with existing neural operators across diverse problems. This training-free approach particularly benefits high-stakes domains like climate modeling and autonomous systems, where error accumulation can compromise prediction reliability.

However, surrogate modeling carries inherent responsibilities. Our method's effectiveness depends critically on accurate PDE residual formulation and evaluation. We emphasize that corrected predictions require careful verification, especially in applications where simulation outputs inform critical decisions. Users should validate results against known benchmarks and consider the method's limitations, particularly regarding computational scaling and discretization errors, when deploying in production environments.

\twocolumn

\end{document}